\documentclass{article}
\usepackage[square, comma, sort, numbers]{natbib}
\usepackage[preprint]{neurips_2024}
\usepackage{verbatim}
\usepackage[most]{tcolorbox}
\usepackage{hyperref}
\usepackage{amsmath}
\usepackage{xcolor}
\usepackage{titlesec}
\definecolor{myorange}{RGB}{0, 70, 0}
\usepackage{neurips_2024}
\usepackage[utf8]{inputenc} 
\usepackage[T1]{fontenc}    
\usepackage{hyperref}       
\usepackage{url}            
\usepackage{booktabs}       
\usepackage{amsfonts}       
\usepackage{nicefrac}       
\usepackage{microtype}      
\usepackage{xcolor}         
\usepackage{graphicx}       
\usepackage{multirow}
\usepackage{multicol}
\usepackage{tikz}
\usepackage{amsmath}
\usepackage{wrapfig}
\usepackage{subcaption}
\usepackage{adjustbox}
\usepackage{colortbl}
\usepackage{booktabs}   
\usepackage{tabularx}   
\usepackage[utf8]{inputenc}

\usepackage{listings}
\usepackage{xcolor}
\usepackage{longtable} 

\lstdefinelanguage{SPARQL}{
    morekeywords={PREFIX,SELECT,DISTINCT,WHERE,FILTER,CONTAINS,LCASE,AND,OPTIONAL,UNION},
    keywordstyle=\color{blue}\bfseries,
    morecomment=[l]{\#},
    commentstyle=\color{green},
}
    
\usepackage{algpseudocode}
\usepackage{algorithm}
\usepackage{amsmath}
\usepackage{xcolor}
\usepackage{pifont}
\usepackage{amsfonts}
\usepackage{hyperref}
\usepackage{enumitem}
\definecolor{deepgreen}{RGB}{0, 100, 0}
\definecolor{deepred}{RGB}{139, 0, 0}

\usepackage[utf8]{inputenc} 
\usepackage[T1]{fontenc}    
\usepackage{hyperref}       
\usepackage{url}            
\usepackage{booktabs}       
\usepackage{amsfonts}       
\usepackage{nicefrac}       
\usepackage{microtype}      
\usepackage{xcolor}         
\usepackage{graphicx}       
\usepackage{wrapfig}        
\usepackage{subcaption}     
\usepackage{makecell}
\usepackage{soul}

\newcommand{\ie}{\textit{i.e.}}
\newcommand{\eg}{\textit{e.g.}}
\definecolor{darkblue}{rgb}{0.0, 0.0, 0.55}
\sethlcolor{red}

\title{Understand What LLM Needs:\\ Dual Preference Alignment for Retrieval-Augmented Generation}

\author{%
    Guanting Dong$^{1}$, 
    Yutao Zhu$^{1}$,
    Chenghao Zhang$^{1}$,
    Zechen Wang$^{2}$,\\
    \textbf{Zhicheng Dou}$^{1}$\thanks{Corresponding author} \ and
    \textbf{Ji-Rong Wen}$^{1}$
    \\
    $^1$Gaoling School of Artificial Intelligence, Renmin University of China.\\
    $^2$School of Artificial Intelligence, Beijing University of Posts and Telecommunications \\
    \small \texttt{\{dongguanting19990611,yutaozhu94\}@gmail.com,} \\
    \small \texttt{dou@ruc.edu.cn}
}

\begin{document}

\maketitle

\begin{abstract}
  Retrieval-augmented generation (RAG) has demonstrated effectiveness in mitigating the hallucination problem of large language models (LLMs). However, the difficulty of aligning the retriever with the diverse LLMs' knowledge preferences inevitably poses an inevitable challenge in developing a reliable RAG system. To address this issue, we propose DPA-RAG, a universal framework designed to align diverse knowledge preferences within RAG systems. Specifically, we initially introduce a preference knowledge construction pipline and incorporate five novel query augmentation strategies to alleviate preference data scarcity. Based on preference data, DPA-RAG accomplishes both external and internal preference alignment: 1) It jointly integrate pair-wise, point-wise, and contrastive preference alignment abilities into the reranker, achieving external preference alignment among RAG components. 2) It further introduces a pre-aligned stage before vanilla Supervised Fine-tuning (SFT), enabling LLMs to implicitly capture knowledge aligned with their reasoning preferences, achieving LLMs' internal alignment. Experimental results across four knowledge-intensive QA datasets demonstrate that DPA-RAG outperforms all baselines and seamlessly integrates both black-box and open-sourced LLM readers. Further qualitative analysis and discussions also provide empirical guidance for achieving reliable RAG systems. Our code is publicly available at \url{https://github.com/dongguanting/DPA-RAG}.
  

\end{abstract}

\section{Introduction}
\label{sec:1}
The emergence of large language models (LLMs)~\citep{ouyang2022training,anil2023palm,openai2024gpt4} has profoundly revolutionized a variety of real-world tasks expressed in natural languages~\citep{chen2021evaluating,longpre2023flan,wei2023chainofthought,luo2023wizardmath,luo2023wizardcoder,yuan2023scaling}. However,  when faced with knowledge-intensive tasks, relying solely on internal knowledge for reasoning may easily expose LLMs to factual inconsistency and hallucination~\citep{bang2023multitask,zhang2023sirens}. To alleviate these issues, researchers use retrieval-augmented technology~\citep{guu2020realm,lewis2021retrievalaugmented} to assist LLMs in integrating relevant external knowledge, providing a promising solution to improve the quality of generated answers~\citep{press-etal-2023-measuring}.

In an ideal Retrieval-Augmented Generation (RAG) system, the goal is to enhance LLMs by incorporating supporting documents that align with their intrinsic knowledge preferences, thus facilitating reasoning.
However, in practical applications, the retriever and the LLM-based reader serve as separate components within the RAG system, each with distinct model architectures, training objectives, and task formats~\citep{bai2023griprank,li2022survey}. These differences often result in documents retrieved by vector similarity failing to meet the specific knowledge demands for LLM reasoning. Moreover, the retrieved documents could even conflict with the self-knowledge of LLMs, potentially disrupting LLMs' original reasoning abilities~\citep{petroni2019language,pmlr-v133-min21a}.

\begin{figure}[!t]
    \centering
    \includegraphics[width=0.98\linewidth]{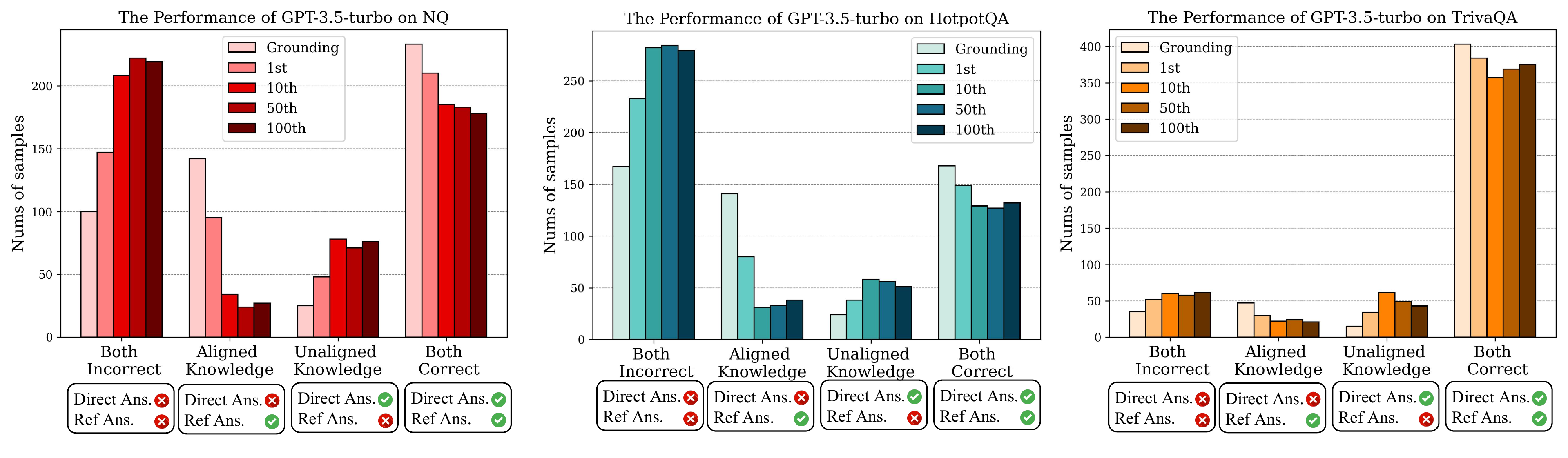}  \vspace{-0.2cm}
    \caption{The results for GPT-3.5 comparing direct responses and answers referencing different retrieved documents (Grounding, 1st, 10th, 50th, 100th) on three QA benchmarks.}
    \label{fig:intro}
\end{figure}

As depicted in Figure~\ref{fig:intro}, we conduct a preliminary analysis on GPT-3.5 across three QA benchmarks, which compare two setups: LLM directly answering question and answering question by referencing different types of retrieved document. We could categorize results into four distinct conditions:
\begin{itemize}[leftmargin=1em]
\item \textit{\textbf{Both
Correct.} The question can be resolved directly by LLMs or through the retrieved document.}
\item \textit{\textbf{Aligned Knowledge.} LLM directly gives the wrong answer, but the retrieved document guide LLM provide right solution.}
\item \textit{\textbf{Unaligned Knowledge.} LLM gives the right answer, but the retrieved document may mislead it.}
\item \textit{\textbf{Both Incorrect.} Neither the retrieved document nor the LLM  can provide an answer correctly.}
\end{itemize}
Then we have the following observations: in the scenario of \emph{``Aligned Knowledge''}, it is notable that documents with low vector similarity (100th) still support LLM in deducing correct answers. Conversely, within the \emph{``Unaligned Knowledge''} scenario, several documents with high vector similarity tend to mislead LLM more than those with lower similarity (\eg, 10th vs 100th). Surprisingly, even some documents that contain relevant grounding information struggle to align with the LLM’s preferences~\citep{lebovitz2021ai}. These results highlight our statement that ``The retrieved documents do not exactly match the knowledge required for LLM reasoning''. Therefore, mitigating the preference gap between the LLM and the retriever emerges as a critical challenge in developing a reliable RAG system.




To address the above limitation, we propose a \underline{D}ual \underline{P}reference \underline{A}lignment for \underline{R}etrieval-\underline{A}ugmented \underline{G}eneration (DPA-RAG), a universal framework designed to align diverse preference knowledge within RAG systems. DPA-RAG consists of three key components: 
(1)~\textbf{Preference Knowledge Construction}: motivated by our preliminary results, we first extract the specific knowledge that significantly affects LLMs' reasoning preferences. Then we introduce five query augmentation strategies and a quality filtering process to synthesize high-quality preference knowledge. 
(2)~\textbf{Reranker-LLM Alignment}: To meet the diverse LLMs' knowledge preferences, we carefully design multi-grained alignment tasks for fine-tuning a preference-aligned reranker. Specifically, we jointly integrate pair-wise, point-wise, and contrastive preference alignment abilities into the reranker via multi-task optimization~\citep{sener2018multi}. By this means, the reranker could provide the necessary knowledge for LLM's inference, achieving external alignment between retriever and LLMs.
(3)~\textbf{LLM Self-Alignment}: To further enable LLMs to concentrate on knowledge aligned with their reasoning preferences, we introduce a pre-aligned phrase prior to the vanilla SFT stage. This stage allows LLMs to capture preference-aligned knowledge from multiple documents, completing the LLM's internal self-alignment. 

To summarize, our contributions are as follows:
\begin{itemize}[leftmargin=1em]


\item Based on a preliminary analysis of GPT-3.5 across three QA benchmarks, we reveal the inherent preference gaps between the retriever and the LLM-based reader in RAG systems.

\item We propose the DPA-RAG, a universal framework designed to align the knowledge preferences of diverse LLMs within RAG systems. DPA-RAG achieves dual preference alignment in two aspects: (1) It jointly integrates multi-grained preference alignment abilities into the reranker, facilitating external alignment across RAG components. (2) It introduces a pre-aligned phrase prior to the standard SFT stage, guiding LLMs to concentrate on the aligned knowledge, thereby unlocking the internal alignment abilities of the LLMs.




\item To overcome the scarcity and limited diversity of preference data, we devise five novel query augmentation strategies and a quality filtering process, aimed at automatically synthesizing high-quality preference data for effectively aligning downstream models.

\item Experimental results on four knowledge-intensive QA datasets demonstrate the effectiveness of DPA-RAG. Further analysis across demensions such as \textit{Model Parameters}, \textit{Preference Alignment}, \textit{Data Quality}, and \textit{Training Strategies} confirm DPA-RAG's role as a plug-and-play solution, providing practical insights for developing reliable RAG systems.


\end{itemize}
\section{Related Work}
 \label{sec:related}
 
\paragraph{Preference Alignment for Large Language Models.}
Traditional Preference alignment (PA) methodologies~\citep{ji2024ai,fang2024domainagnostic,wang2023aligning,jiang-etal-2023-structgpt} are designed to tailor pre-trained language models to reflect human preferences. Recently, a series of works have relied on reinforcement learning (RL)~\citep{DBLP:journals/corr/ppo} to align LLMs with human preferences~\citep{ouyang2022training}.
Owing to the sensitivity of RL's parameters and the complex process of reward modeling, research works~\citep{liu2023alignment-add6,DBLP:journals/corr/alignment-add2,DBLP:journals/corr/alignment-add4,nathani2023alignment,DBLP:journals/corr/alignment-add1,DBLP:journals/corr/alignment-add5,DBLP:journals/corr/alignment-add3,DBLP:journals/corr/slic-hf,song2024preference} represented by DPO~\citep{rafailov2023direct} further tried to optimize the loss function and reward scoring mechanism for pruning. However, depending on annotations from humans or expert models still increases the alignment cost. To construct reliable RAG systems, a branch of studies~\citep{shi2023replug,bonifacio2022inpars,jeronymo2023inparsv2} aims to align the retriever with supervision signals generated by LLMs, showcasing remarkable alignment potential.  
Conversely, other studies attempt to improve the alignment abilities of RAG systems by implementing a multi-round retrieval paradigm~\citep{jiang2023active,yao2023react,ren2023investigating,zhou2024metacognitive,trivedi2023interleaving,wang2024llms} and filtering out noise from the training corpus~\citep{zhang2024raft,wang2023learning,zhang2024knowledgeable,jin2024bider, wang2024rat}. These approaches, however, often suffer from a lack of multi-level alignments, which limits their ability to adapt to the diverse knowledge preferences of LLMs. In our paper, we introduce DPA-RAG, a system that bridges this gap by aligning the retriever to adapt to the diverse knowledge preferences of LLMs without relying on external expert annotations.

\paragraph{Reranking Techniques for Retrieval Augmented Generation.}
In the RAG system, the reranker is designed to rank a list of retrieved documents to accurately meet LLMs' demands. A series of sentence transformer models~\citep{reimers2019sentencebert,bge_embedding, khattab2020colbert,nogueira2019multistage} have achieved excellent fine-grained ranking by better aligning the representations between queries and documents.
With the rapid development of prompt learning~\citep{liu2023pre}, point-wise generative re-ranking frameworks~\citep{nogueira-etal-2020-document,ju2022texttotext,pradeep2021expandomonoduo,zhuang2022rankt5} have transformed traditional discriminative tasks into a Seq2seq paradigm, showcasing promising initial alignment abilities.
The recent development and application of LLMs have introduced innovative pair-wise and list-wise rerankers, such as RankGPT~\citep{sun2023chatgpt}, PRP~\citep{qin2024large}, LRL~\citep{ma2023zero} and RankLLaMA~\citep{ma2023finetuning}. These models have brought multi-perspectives in addressing the fine-grained re-ranking problem.
Moreover, in response to the unique preferences of different users, various methods~\citep{pei2019personalized,li2022pear,saadfalcon2023udapdr,ma2023large,shi2022xricl} have been developed to achieve personalized user sorting, yielding significant results in aligning with industrial scenarios.
These advancements inspire us to distill the preferences of LLMs into the reranker, facilitating effective alignment between the RAG system's components.

\section{Methodology}

 \label{sec:method}
To address the misalignment between different components of retrieval-augmented generation (RAG) and improve overall generation performance, we propose the DPA-RAG framework, which is illustrated in Figure~\ref{fig:main}. In general, DPA-RAG improves traditional RAG architecture in two main aspects: (1) we fine-tune a preference-aligned reranker between the retriever and the LLM to selectively filter out knowledge that aligns with LLMs' knowledge preferences ($\S$\ref{sec:33}); and (2) we design a self-alignment mechanism that fine-tunes LLMs to better recognize and utilize knowledge consistent with their reasoning preferences ($\S$\ref{sec:34}). To acquire the LLM's preference knowledge, we devise a three-step construction method, motivated by our preliminary analysis of how different types of retrieved documents affect RAG performance ($\S$\ref{sec:32}). Below, we will first introduce the task definition ($\S$\ref{sec:31}) and then we delve into the specifics of our approach.

\subsection{Task Definition}
\label{sec:31}
Compared to standard text generation, RAG often follows a \textit{retrieve-then-read} paradigm~\citep{lewis2021retrievalaugmented}, where an additional retriever is introduced to collect external knowledge and enhance the generation process. This architecture involves constructing a \textit{query} $q$ to reflect the information needs of the generation. For example, in question-answering systems, the input question is often used as the query. Given the query $q$, the retriever $R$ returns relevant documents from a corpus $D_{q}=\{d_i\}^{N}_{i=1}$ with $N$ documents. The relevance between document $d$ and query $q$ can be measured by various methods. In this work, we employ a dense retriever that utilizes dual encoders to obtain hidden representations for both the query and the documents. The relevance score is then calculated by computing the dot-product similarity between these representations, enabling the retrieval of the top-$k$ documents $D_{\text{retrieve}}$:
\begin{align}
D_{\text{retrieve}} = {\text{argtop-}k} \left[ E_{\text{d}} (d_i)^{\top} \cdot E_{\text{q}}(q) \mid i = \{1 \ldots N\} \right].
\end{align}
While the retrieved documents are relevant to the query, they may not necessarily contain the knowledge required by the LLMs. Therefore, in this study, we introduce a reranker $E_r$ to rerank $D_{\text{retrieve}}$ and filter out the documents $D_{\text{rerank}}$, which include only those documents aligned with the LLMs' preferences \ie, $D_{\text{rerank}} = E_{r}(q, D_{\text{retrieve}})$.
Finally, the LLMs read from the reranked documents and generate the target text based on the query: 
\begin{align}
    y = \text{LLM}(q,D_{\text{rerank}}) = {\log{{P}_{\theta}\left(q, D_{\text{rerank}}\right)}},
\label{eq::task}
\end{align}
where ${P}_{\theta}$ represents the LLM's generation probability distribution. 

Recognizing that LLMs might struggle to effectively utilize retrieved knowledge, we also design a self-alignment mechanism to optimize $\theta$ for RAG tasks.

\subsection{Preference Knowledge Construction}
\label{sec:32}
\begin{figure}[t]
    \centering
    \resizebox{\textwidth}{!}{
    \includegraphics{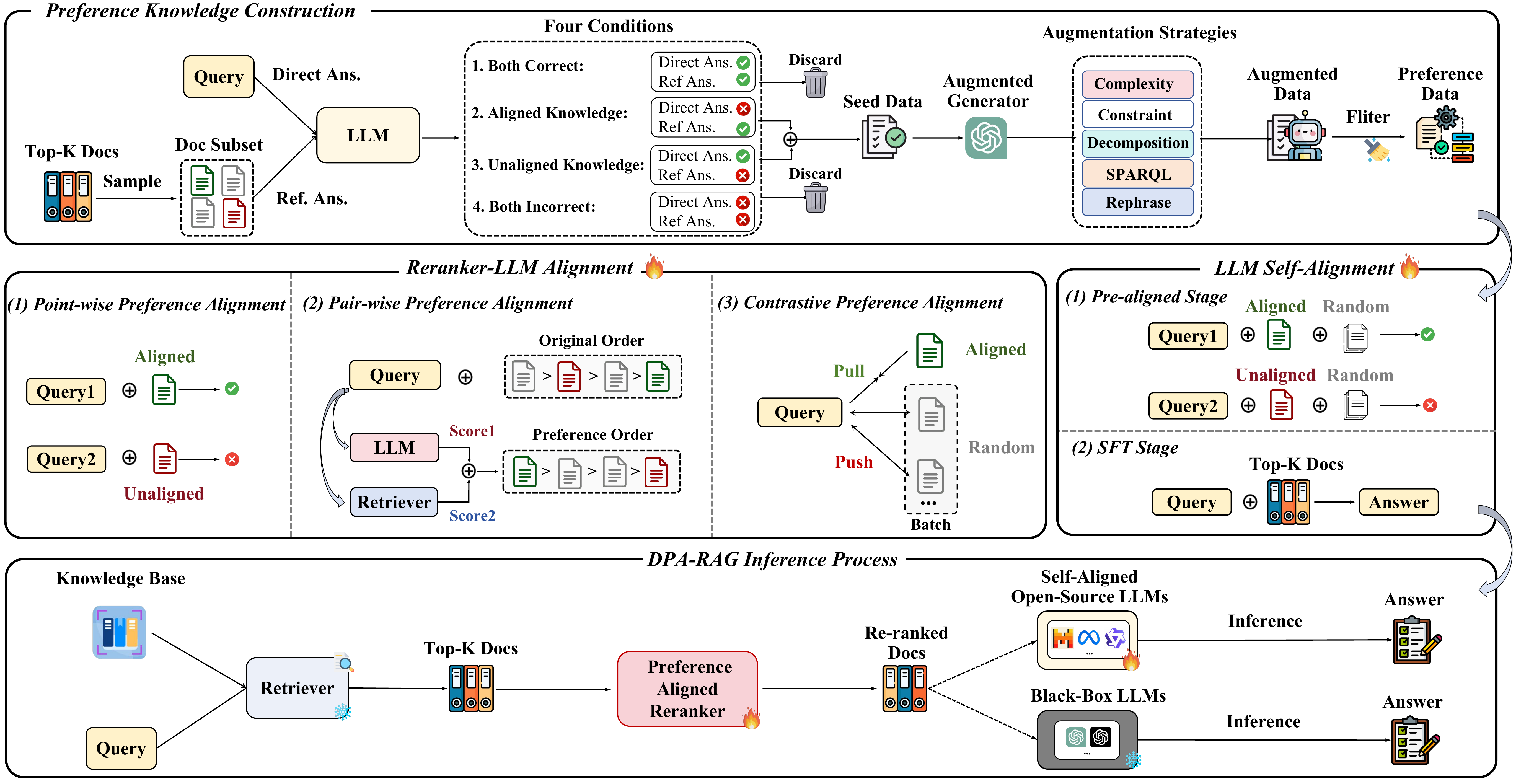}
    }
    \caption{The overall framework of DPA-RAG. The upper part shows the pipeline for preference knowledge construction. The middle part displays the task format for dual preference alignment. The bottom part illustrates the inference process of DPA-RAG.}
    \label{fig:main}
\end{figure}

To mitigate the misalignment between different RAG components, a critical step is to collect data that reflects LLMs' knowledge preferences. Therefore, we design a three-step method to gradually mine, augment, and filter out high-quality preference knowledge of LLM, which is shown in Figure~\ref{fig:main}.

\paragraph{Preference Knowledge Extraction.}
To align with LLMs' knowledge preferences, it is essential to identify the specific knowledge that can bring performance gains or harms during the model's inference process. 
Motivated by the preliminary analysis in Figure \ref{fig:intro}, given the training set $ \widetilde{D}_{\text{train}} = \{q_{i},D_{q_{i}}, y_{q_{i}} \}_{i=1}^{N_{\text{train}}}$, where each sample includes a query $q_i$, top-$k$ retrieved documents $D_{q_{i}}=\{d_{i}\}_{i=1}^{k}$ and an answer $y_{q_{i}}$. 
We guide LLMs to directly answer questions or response by referencing different types of documents, aiming to filter out samples from $\widetilde{D}_{\text{train}}$ that reflects LLMs’ knowledge preferences. 

To ensure the distinctiveness among these documents, we hierarchically sample four documents from $D_{q_{i}}$ to construct the document subset $D_{q_{i}}^{\text{sub}} = \{d_{i} | i = 1, 25, 50, 100\}$ for each query, as shown in the upper part of Figure \ref{fig:main}. Consequently, we also categorize the results of LLMs into \emph{Both Correct''}, \emph{Both Incorrect''}, \emph{Aligned Knowledge''}, and \emph{Unaligned Knowledge''}. From $\widetilde{D}{\text{train}}$, we selectively extract samples whose document subsets $D_{q_{n}}^{\text{sub}}$ contain at least one document labeled \emph{Aligned Knowledge''} or \emph{Unaligned Knowledge''}. This allows us to obtain the preference dataset $\widetilde{D}_{\text{pref}} = \{q{i}, D_{q_{i}}^{\text{sub}}, Y_{i}^{\text{sub}}\}_{i = 1}^{N}$, where $Y_{i}^{\text{sub}} = \{y_{i} | i = 1, 25, 50, 100\}$ denotes the preference labels of $D_{q_{i}}^{\text{sub}}$, corresponding to the four distinct categories.

The motivation behind this selection process is that documents labeled as \emph{Aligned Knowledge''} or \emph{Unaligned Knowledge''} provide the LLM with a clear positive or negative impact during reasoning. Due to the difficulty in distinguishing the role of retrieved documents labeled \emph{Both Correct''} or \emph{Both Incorrect''}, we choose to discard them.

\paragraph{Diverse Query Augmentation.}
Upon obtaining $\widetilde{D}_{\text{pref}}$ which reflects LLM's preferences, its scarcity $\widetilde{D}_{\text{pref}}$ (only 20\% of $\widetilde{D}_{\text{train}}$) still poses an obstacle for fine-tuning high-quality models. More critically, the sparsity of preference data results in limited data patterns, reducing both the diversity and complexity of the dataset~\citep{liu2024makes,zeng2024automatic}. To address these limitations, we are inspired by several augmentation methods~\citep{luo2023wizardmath, luo2023wizardcoder, yu2023metamath, yuan2023scaling, li2023query} and specifically design five novel query augmentation strategies for the RAG system as follows\footnote{Detailed information on the different augmentation strategies can be found in Appendix \ref{app:aug}}:
\begin{itemize}[leftmargin=1em]
\item \textit{\textbf{Rephrasing.} Rephrase the original query with
the same intention.}
\item \textit{\textbf{Complexity.} Increase the semantic complexity of the original query.}
\item \textit{\textbf{Decomposition.} Decompose the original query into several sub-problems.}
\item \textit{\textbf{Constraint.} Add more conditional
and constrained statements to the original query.}
\item \textit{\textbf{SPARQL.} Rewrite the original query based
on the SPARQL syntax and generate it directly.}
\end{itemize}
We utilize GPT-3.5-turbo to generate different augmented datasets $\{\widetilde{D}_{r_i}\}_{i=1}^{n}$, and then merge them with original dataset $\widetilde{D}_{\text{ori}}$, which can be formulated as $\widetilde{D}_{\text{pref}}^{\text{ori}} = \widetilde{D}_{\text{pref}}^{\text{ori}} \cup ( \cup_{i=1}^{n} \widetilde{D}_{r_i} )$.

To control the augmented data's quality, we introduce a quality filtering procedure by a natural language inference (NLI) model. Given the original query $q$ as the \emph{``premise''} and the augmented query $q_{\text{aug}}$ as the \emph{``hypothesis''}, the NLI model seeks to determine the semantic relationship between the two queries. The relation can be categorized as \emph{entailment}, \emph{contradiction}, or \emph{neutral}, as follows:
\begin{equation}
p_{\theta}(\cdot \mid q, q_{\text{aug}})=\operatorname{softmax}\left({\operatorname{score}}_{\theta}(q, q_{\text{aug}})\right),
\end{equation}
where $\operatorname{score}_{\theta}: \mathbb{R}^{k \times \ell_{q}} \times \mathbb{R}^{k \times \ell_{q_{aug}}} \rightarrow \mathbb{R}^{3}$ is a scoring function dependent on the model's parameters $\theta$. To maintain intent consistency between the original and augmented datasets, we exclude any augmented data labeled as "contradiction" (approximately 20\%).


\subsection{Reranker-LLM Alignment}
\label{sec:33}

After obtaining $D_{\text{pref}}$, we introduce multi-grained preference alignment tasks to jointly fine-tune a reranker, aiming to filter retrieved knowledge that aligns with LLM preferences.

\paragraph{Point-wise Preference Alignment.}
Distinguishing beneficial or harmful knowledge of LLMs is essential for aligning their preferences. Hence, from each sample $\{q_{i},D_{q_{i}}^{\text{sub}}, Y_{i}^{\text{sub}}\}\sim\widetilde{D}_{\text{pref}}$, we can further extract one sub-sample $\{q_{i},d_{i}, y_{i}\}$ where $y_i$ is labeled as \emph{``Aligned Knowledge''} or \emph{``Unaligned Knowledge''}.
As shown in Figure \ref{fig:main}, we use $\{q_{i},d_{i}, y_{i}\}_{i=1}^{N}$ to fine-tune the Reranker model $E_{r}(\theta)$ with binary cross-entropy loss~\citep{shannon1948mathematical}, achieving a point-wise preference alignment:
\begin{equation}
\mathcal{L}_{\text{point}} = -\frac{1}{N}\sum_{i=1}^{N} \ [y_i\log(p_{\theta}(q_{i},d_{i})) + (1 - y_i)\log(1 - p_{\theta}(q_{i},d_{i}))],
\end{equation}
where $y_i$ is label (Postive / Negative) for judging the $d_i$ is aligned or unaligned knowledge.

\paragraph{Pair-wise Preference Alignment.}
Since point-wise alignment empowers the reranker to identify LLM's favored knowledge, enhancing the reranker to prioritize this preferred knowledge presents a new challenge. Therefore, we propose a pair-wise preference ranking task for fine-grained alignment. In detail, given $\{q_{i},D_{q_{i}}^{\text{sub}}, y_{i}^{\text{sub}}\}\sim\widetilde{D}_{\text{pref}}$, we derive an order $  \{o_{i}\}_{i=1}^{K}$ of the documents subset $D_{q_{i}}^{\text{sub}} = \{d_i\}_{i=1}^{K}$ based on the initial similarity scores from the retriever.

Our idea is elegantly simple: we leverage the LLM within the RAG system as a preference reward model $r_{\theta}$ to score documents, eliminating the need for external experts.  To mitigate bias from relying solely on LLM-generated preference scores~\citep{zhuang2023open}, we calculate the preference score $s_i$ for each query by weighting both the LLM preference score $r_{\theta}$ and the original similarity score $s_{R}(\cdot)$ from the retriever:
\begin{equation}
s_i = a \cdot r_{\theta}(q, d_i) + (1-a) \cdot s_{R}(q, d_i),
\end{equation}
$s_i$ denotes the preference score of the $i$-th retrieved document. We then sort the documents according to these preference scores to obtain the LLM's knowledge preference order $\{\hat{o_{i}}\}_{i=1}^{K}$. Subsequently, we integrate the preference order into the reranker using RLHF loss~\citep{stiennon2022learning, ouyang2022training}:
\begin{equation}
\mathcal{L}_{\text{pair}}=-\frac{1}{C_{k}^2} \ \mathbb{E}_{\left(q,d_{w},d_{l}, y_{w}, y_{l}\right) \sim \widetilde{D}_{\text{pref}}} \ \left[\log \left(\sigma\left(p_{\theta}\left(q,d_{w}, y_{w}\right)-p_{\theta}\left(q,d_{l}, y_{l}\right)\right)\right)\right],
\end{equation}
where $y_w$ and $y_l$ represent the labels for documents $d_w$ and $d_l$, corresponding to \emph{“winner”} or \emph{“loser”} in the preference order ${\hat{o_{i}}}{i=1}^{K}$. $p{\theta}$ denotes the logits of the output.\footnote{An in-depth discussion on scoring mechanisms for different LLMs can be found in Appendix~\ref{app:score}.}


\paragraph{Contrastive Preference Alignment.}
To align query representations with the LLM's preferred knowledge, we employ contrastive learning~\citep{oord2019representation, bachman2019learning} to fine-tune our reranker, thereby preventing the LLM from being misled by highly similar but unaligned knowledge. Unlike previous pairwise approaches~\citep{rafailov2023direct}, our $\widetilde{D}{\text{pref}}$ dataset associates each query with multiple documents, rather than a single positive or negative example. Considering this one-to-N scenario, we employ Supervised Contrastive Learning (SCL)~\citep{khosla2021supervised} to fully leverage $\widetilde{D}{\text{pref}}$. In our task, the query serves as an anchor point $h_{q}$. Aligned documents are treated as positive samples $h_{p}$, while documents randomly sampled from other instances in the batch act as negative samples $h_{n}$. As shown in Figure \ref{fig:main}, SCL seeks to reduce the distance of queries and positive samples $h_{p}$, while increasing the distance from negative samples $h_{n}$ in the semantic space. The loss $\mathcal{L}_{\text{CPA}}$ is formulated as follows:
\begin{equation}
\mathcal{L}_{\text{CPA}}= -\sum_{i=1}^{N_{t}}\frac{1}{N_{y_{i}}-1} \sum_{j=1}^{N_{t}} \mathbf{1}_{i \neq j} \mathbf{1}_{y_{i}=y_{j}} \log \frac{\exp (h_q \cdot h_p / \tau)}{\sum_{k=1}^{N_{t}} \mathbf{1}_{i \neq k} \exp (h_q \cdot h_n / \tau)},
\end{equation}
$N_{t}$ is the nums of samples in each batch. $N_{y_i}$ denotes samples in the batch with same label as $y_i$. $\tau$ is a temperature parameter. \textbf{1} is an indicator.


\paragraph{Multi-task Optimization.}
Optimizing multi-grained preference tasks via Multi-task Learning (MTL)~\citep{caruana1997multitask,romera2013multilinear} offers a efficient way for fine-tuning the reranker. However, learning tasks jointly may further introduce potential bias and conflicts~\citep{luong2015multi}. To tackle this challenge, we employ the MGDA-UB~\citep{sener2018multi}, aiming to dynamically find a pareto optimal~\citep{lin2019pareto} solution for balancing multi-task optimization.
By utilizing MGDA-UB to optimize the MTL weights $\{c^t\}_{t=1}^{T}$ for $T$ tasks. We finally obtain our multi-grained alignment loss function as:
\begin{equation}
\mathcal{L}_{\text{total}} = c^{1}  \mathcal{L}_{\text{point}}
+ c^{2}  \mathcal{L}_{\text{pair}} + c^{3} \mathcal{L}_{\text{CPA}}
\end{equation}

\subsection{LLM Self-Alignment}
\label{sec:34}

After initially aligning the preferences between external RAG components, in this section, we focus on guiding LLMs to emphasize aligned knowledge during the reasoning process to achieve internal alignment. Inspired by several pre-alignment works~\citep{liu2020self,wang2024dolphcoder}, we introduce a pre-aligned stage to assist LLMs in implicitly identifying the knowledge crucial for reasoning~\citep{jin2024bider}.


\textbf{Pre-aligned Stage.} As illustrated in Figure \ref{fig:main}, for each sample $\{q_{i},D_{q_{i}}^{\text{sub}}, Y_{i}^{\text{sub}}\}\sim\widetilde{D}_{\text{pref}}$, we randomly select one document $d_{q}$ labeled \emph{``Aligned Knowledge''} or \emph{``Unaligned Knowledge''} from $D_{q_{i}}^{\text{sub}}$, along with $k-1$ random documents from the retrieved corpus $D=\{d_i\}^{N}_{i=1}$. This selection process constructs a top-$k$ document set $D_{\text{align}} = \{d_{q}, d_{{\text{rand}}_1}, \ldots, d_{{\text{rand}}_{k-1}}\}$ for each query $q$. Then we perform the following training objective with task specific template\footnote{The document $d_q$ is placed at a random position among $k$ documents.}:
\begin{equation}
\label{eq14}
    \mathcal{L}\left(\theta\right) = {\sum\limits_{(q_{n},D_{q}, y_{n}) \in D_{\text{pref}} }{\log{{P}_{\theta}\left(y_n | \operatorname{prompt}(q_{n},D_{\text{align}}) \right)}}},
\end{equation}
\begin{description}
    \item \small\textbf{Prompt}: \colorbox{gray!8}{\parbox{0.95\linewidth}{\small
    Given the documents $\{D_{\text{align}}= (d_{q}, d_{{\text{rand}}_1}, \ldots, d_{{\text{rand}}_{k-1}})\}$. Answer the following question based on the given information or your internal knowledge with few words without the source. Query: $\{q\}$. \\
    \textcolor{darkblue}{[Judgement]: document-$\{i_{d_{q}}\}$ is Positive or Negative knowledge for answering question.}}}
\end{description}




where $ \log {P}(\cdot )$ denote probability distribution of LLM's output. $\theta$ denotes model parameters. $\{i_{d_{q}}\}$ represents the position of the preference document. LLMs will implicitly learn the ability to capture self-preferred knowledge from top-$k$ documents by distinguishing $y \in {\text\{positive, negative\}}$ during pre-aligned task.


\textbf{Supervied Fine-tuning Stage.} Following the pre-aligned task, we load pre-trained parameters and perform subsequent Supervised Fine-tuning (SFT) for QA tasks using the same objective described in Equation (\ref{eq14}). We utilize the traditional QA format training set $\widetilde{D}_{\text{train}} = \{q_{i}, D_{q_{i}}, y_{q_{i}}\}_{i=1}^{N_{\text{train}}}$. Moreover, we merge five augmented datasets $\{\widetilde{D}
_{r_i}\}_{i=1}^{5}$ with $\widetilde{D}_{\text{train}}$. Using the preference-aligned reranker $E_r$, we reorder the documents and filter out the top-k documents as described in Equation (\ref{eq:rerank}), forming the final training set $\widetilde{D}_{\text{train}}^{\text{rank}} = \{q_{i}, D_{q_i}^{\text{rank}}, y_{q_{i}}\}_{i=1}^{N_{\text{train}}}$ of SFT stage.
\begin{equation}
\label{eq:rerank}
D_{q_i}^{\text{rank}} = {\text{argtop-}k} \left[ E_{r}(q_i, D_{q_i})\right]
\end{equation}
The preference knowledge identification capability developed during the pre-alignment stage enables LLMs to focus more effectively on aligned knowledge during the SFT stage, thereby enhancing their internal alignment potential. The prompt template for SFT stage is as follows:
\begin{description}
    \item \small\textbf{Prompt}: \colorbox{gray!8}{\parbox{0.95\linewidth}{\small
    Given the documents $\{\text{Top-K Docs: } D_{q}^{\text{rank}}\}$. Answer the following question based on the given information or your internal knowledge with few words without the source. Query:$\{q\}$.}}
\end{description}


\section{Experiments}

\subsection{Datasets and Metrics}
\begin{wraptable}{r}{0.48\textwidth}
\vspace{-0.5cm}
    \tiny
    \renewcommand{\arraystretch}{0.9} 
    \setlength{\tabcolsep}{0.5mm} 
  \centering
    \caption{Statistics for the QA datasets.}
  \resizebox{0.48\textwidth}{!}{
  \begin{tabular}{l|>{\centering\arraybackslash}p{1cm}>{\centering\arraybackslash}p{1cm}>{\centering\arraybackslash}p{1cm}}
    \toprule
    \multirow{2}{*}{\textbf{Dataset}} & \multicolumn{3}{c}{\textbf{\# Examples} (thousands)} \\
    & \textbf{Train} & \textbf{Dev} & \textbf{Test} \\
    \midrule
    NQ           & 79.2 & 8.7 & 3.6  \\
    TriviaQA     & 78.8 & 8.8 & 11.3 \\
    HotpotQA     & 88.9 & 5.6 & 5.6  \\
    WebQSP & 2.84  & 0.25   & 1.6  \\
    \bottomrule
  \end{tabular}
  }
  \label{data-statistic}
  \vspace{-0.2cm}
\end{wraptable}
\label{sec::Datasets and Metrics}
We select four question answering (QA) datasets covering three types, including \textbf{(1) Open-Domain QA}, represented by NaturalQuestions (NQ)~\citep{NQ} and TriviaQA (TQA)~\citep{TriviaQA}; \textbf{(2) Multi-Hop QA}, represented by HotpotQA (HQA)~\citep{HotpotQA}; and \textbf{(3) Knowledge Base QA}, represented by WebQuestionsSP (WebQSP)~\citep{webqsp}. Table~\ref{data-statistic} illustrate the statistics of them. For evaluation metrics, we use Hit@1 for the accuracy of the top-ranked response and F1 score to assess the quality and similarity to the ground-truth. More details of the experimental setup are listed in Appendix~\ref{app:setup}.

\begin{table}[!t]
    \centering
    \small
        \renewcommand{\arraystretch}{1.1} 
    \setlength\tabcolsep{1pt}
    \caption{The main results of DPA-RAG and different kinds of baselines on four QA benchmarks.}
    \resizebox{0.95\textwidth}{!}{
    \begin{tabular}{llcccccccc}
    \toprule
    \multirow{2}{*}{\textbf{Method}} & \multirow{2}{*}{\textbf{Reader}} & \multicolumn{2}{c}{\textbf{NQ}} & \multicolumn{2}{c}{\textbf{Trivia-QA}} & \multicolumn{2}{c}{\textbf{Hotpot-QA}} & \multicolumn{2}{c}{\textbf{WebQSP}} \\
    \cmidrule(r){3-4}\cmidrule(lr){5-6}\cmidrule(lr){7-8}\cmidrule(l){9-10}
    & &  \textbf{Hit@1} & \textbf{F1} & \textbf{Hit@1} & \textbf{F1} & \textbf{Hit@1} & \textbf{F1} & \textbf{Hit@1} & \textbf{F1} \\
    \midrule
    \multicolumn{9}{c}{Traditional RAG with DPR} \\
    \midrule
    RAG~\citep{DBLP:conf/nips/instructgpt}  & GPT-3.5   & 47.47 & 47.99 & 75.04 & 74.13 & 26.28 & 32.84 & 67.97 & 63.33 \\
    RAG~\citep{achiam2023gpt}                & GPT-4      & 54.04 & 51.19 & 79.98 & 76.85 & 28.46 & 33.87 & 71.30 & 67.20 \\
    RAG~\citep{touvron2023LLaMA}         & LLaMA2-7B  & 50.94 & 54.76 & 63.90 & 63.80 & 31.40 & 38.90 & 68.52 & 64.22 \\
    RAG~\citep{touvron2023LLaMA}        & LLaMA2-13B & 56.60 & 60.60 & 70.43 & 71.32 & 36.31 & 45.23 & 76.39 & 78.63 \\
    RAG~\citep{LLaMA3}         & LLaMA3-8B  & 54.81    & 58.33    & 69.54  & 71.21 &34.28     & 42.29      & 72.82     & 73.94        \\
    RAG~\citep{qwen}         & Qwen2-7B  & 52.01   & 56.13    & 63.88  & 66.52 &31.39  & 39.70   & 75.98    & 77.82      \\
    \midrule
    \multicolumn{9}{c}{RAG with DPR \& Reranker} \\
    \midrule
    RAG+RankGPT~\citep{sun2023chatgpt}                 & LLaMA2-7B  & 47.81 & 52.37 & 59.05 & 56.39 & 28.32 & 37.06 & 66.32 & 62.22 \\
    RAG+LRL~\citep{ma2023zero}                         & LLaMA2-7B  & 48.09 & 53.06 & 60.33 & 56.86 & 29.13 & 37.81 & 67.43 & 63.44 \\
    RAG+PRP~\citep{qin2024large}                       & LLaMA2-7B  & 51.91 & 56.17 & 62.28 & 57.98 & 31.90 & 40.87 & 68.54 & 64.08 \\
    RAG+RankLLaMA~\citep{ma2023finetuning}             & LLaMA2-7B  & 52.18 & 56.62 & 62.34 & 58.05 & 32.31 & 41.39 & 69.11 & 65.70 \\
    RAG+BGE~\citep{bge_embedding}                       & LLaMA2-7B  & 52.43 & 56.92 & 62.70 & 57.58 & 32.53 & 41.73 & 70.20 & 68.80 \\
    RAG+BCEmbedding~\citep{youdao_bcembedding_2023}            & LLaMA2-7B  & 49.91 & 53.19 & 61.93 & 57.67 & 31.52 & 40.59 & 68.20 & 65.40 \\
    RAG+ColBERTv2~\citep{santhanam2022colbertv2}       & LLaMA2-7B  & 51.49 & 56.02 & 62.34 & 58.16 & 31.72 & 40.79 & 69.70 & 66.90 \\
    \midrule
    \multicolumn{9}{c}{{Preference-aligned Methods for RAG}} \\
    \midrule
    KnowPAT~\citep{zhang2024knowledgeable}         & LLaMA2-7B  & 51.42 & 54.82 & 63.20 & 65.20 & 29.00 & 37.40 & 68.73 & 65.31 \\
    REPLUG~\citep{shi2023replug}                   & GPT-3.5    & 49.67 & 50.58 & 75.67 & 75.34 & 27.30 & 34.30 & 69.59 & 66.22 \\
    RA-Judgement~\citep{ren2023investigating}      & GPT-3.5    & 48.52 & 50.18 & 76.21 & 76.58 & 26.50 & 32.81 & 66.07 & 68.32     \\
    RRHF~\citep{yuan2023rrhf}                      & LLaMA2-7B  & 50.11 & 52.01 & 62.50 & 60.20 & 28.16 & 35.40 & 66.90 & 63.10 \\
    RAFT~\citep{zhang2024raft}                     & LLaMA2-7B  & 50.24 & 53.86 & 60.10 & 57.40 & 30.20 & 35.80 & -     & -     \\
    FILCO~\citep{wang2023learning}                 & LLaMA2-7B  & 52.71 & 55.32 & 67.30 & 67.80 & 32.70 & 40.80 & 69.96 & 68.34 \\
    \midrule
    \multicolumn{9}{c}{{Our Method: DPA-RAG}} \\
    \midrule
    DPA-RAG                                & GPT-3.5    & 51.60 \tiny{\textcolor{myorange}{(+4.13)}} & 52.80 \tiny{\textcolor{myorange}{(+4.81)}} & 78.65 \tiny{\textcolor{myorange}{(+3.61)}} & 77.05 \tiny{\textcolor{myorange}{(+2.92)}} & 28.42 \tiny{\textcolor{myorange}{(+2.14)}} & 36.12 \tiny{\textcolor{myorange}{(+3.28)}} & 71.80 \tiny{\textcolor{myorange}{(+3.83)}} & 69.20 \tiny{\textcolor{myorange}{(+5.87)}} \\
    DPA-RAG                                  & GPT-4      & 56.45 \tiny{\textcolor{myorange}{(+2.41)}} & 53.28 \tiny{\textcolor{myorange}{(+2.09)}} & 84.41 \tiny{\textcolor{myorange}{(+4.43)}} & 80.08 \tiny{\textcolor{myorange}{(+3.23)}} & 33.79 \tiny{\textcolor{myorange}{(+5.33)}} & 37.67 \tiny{\textcolor{myorange}{(+3.80)}} & 73.12 \tiny{\textcolor{myorange}{(+1.82)}} & 74.83 \tiny{\textcolor{myorange}{(+7.63)}} \\
    DPA-RAG                              & LLaMA2-7B  & 56.03 \tiny{\textcolor{myorange}{(+5.09)}} & 60.19 \tiny{\textcolor{myorange}{(+5.43)}} & 70.16 \tiny{\textcolor{myorange}{(+6.26)}} & 70.29 \tiny{\textcolor{myorange}{(+6.49)}} & 35.23 \tiny{\textcolor{myorange}{(+3.83)}} & 43.34 \tiny{\textcolor{myorange}{(+4.44)}} & 72.40 \tiny{\textcolor{myorange}{(+3.88)}} & 71.80 \tiny{\textcolor{myorange}{(+7.58)}} \\
    DPA-RAG                             & LLaMA2-13B & 59.19 \tiny{\textcolor{myorange}{(+2.59)}} & 62.97 \tiny{\textcolor{myorange}{(+2.37)}} & 74.18 \tiny{\textcolor{myorange}{(+3.75)}} & 75.53 \tiny{\textcolor{myorange}{(+4.31)}} & 41.07 \tiny{\textcolor{myorange}{(+4.76)}} & 49.60 \tiny{\textcolor{myorange}{(+4.37)}} & 80.28 \tiny{\textcolor{myorange}{(+3.89)}} & 81.74 \tiny{\textcolor{myorange}{(+3.11)}} \\
    DPA-RAG                              & LLaMA3-8B  & 57.43\tiny{\textcolor{myorange}{(+2.62)}}    & 61.02 \tiny{\textcolor{myorange}{(+2.69)}}     & 72.04\tiny{\textcolor{myorange}{(+2.50)}}      & 73.58 \tiny{\textcolor{myorange}{(+2.37)}}    & 36.01 \tiny{\textcolor{myorange}{(+1.73)}}     & 44.32  \tiny{\textcolor{myorange}{(+2.03)}}    & 74.26 \tiny{\textcolor{myorange}{(+1.44)}}     & 76.11 \tiny{\textcolor{myorange}{(+2.17)}}     \\
    DPA-RAG                              & Qwen2-7B  & 54.66\tiny{\textcolor{myorange}{(+2.65)}}    & 58.84 \tiny{\textcolor{myorange}{(+2.71)}}     & 68.58\tiny{\textcolor{myorange}{(+4.70)}}      & 70.26 \tiny{\textcolor{myorange}{(+3.74)}}    & 34.56 \tiny{\textcolor{myorange}{(+2.87)}}     & 42.47 \tiny{\textcolor{myorange}{(+2.77)}}    & 78.66 \tiny{\textcolor{myorange}{(+2.68)}}     & 80.53 \tiny{\textcolor{myorange}{(+2.71)}}     \\
    \bottomrule
    \end{tabular}
    }
    \label{main-results}
\end{table}

\subsection{Main Results}
The experimental results are shown in Table~\ref{main-results}. In general, our DPA-RAG significantly outperforms all baselines across four datasets in different setups. This clearly highlights the superiority of our approach. We further have the following observations: 

(1) Compared to traditional RAG baselines, DPA-RAG (LLaMA2-7B) shows a remarkable performance improvement (over 5\%) across all four datasets. More importantly, this improvement is consistent across various models, including LLaMA2-13B, Qwen2-7B, LLaMA3-8B, GPT-3.5, and GPT-4. This indicates the broad applicability and generalizability of our method.

(2) For reranker-based methods, we find that smaller rerankers such as BGE and ColBERTv2 can achieve comparable or even better performance than LLM-based rerankers. This result validates our motivation of using BGE as the alignment backbone, as it combines efficiency with effectiveness.

(3) Among preference-aligned methods, DPA-RAG outperforms direct alignment methods (\ie, REPLUG and RA-Judgement), which rely on logits. This emphasizes the value of implementing multi-grained alignments within our framework. Surprisingly, Filco, which employs data filtering, shows robust alignment capabilities, confirming that unaligned knowledge exists in training corpora. This observation highlights again the importance of our preference optimization at the data level, ensuring that the retrieved and used knowledge is highly relevant and aligned with the LLM’s needs.


\begin{wraptable}{r}{0.45\textwidth}
    \vspace{-1.5em}
    \centering
         \renewcommand{\arraystretch}{1.2} 
    \caption{Ablation study on NQ and TQA.}
    \scalebox{0.6}{
    \begin{tabular}{lcccc}
    \toprule
     \textbf{Method}   & \multicolumn{2}{c}{\textbf{NQ}} & \multicolumn{2}{c}{\textbf{TQA}} \\
    \cmidrule(r){2-3}\cmidrule(l){4-5}
        & \textbf{Hits@1} & \textbf{F1} & \textbf{Hits@1} & \textbf{F1} \\
    \midrule
    LLaMA2-7B RAG & 50.94 & 54.76 & 63.90 & 63.80 \\
    LLaMA2-7B DPA-RAG & 56.03 & 60.19 & 70.16 & 70.29 \\
    \rowcolor{blue!10} \quad \textit{w/o} PA-Rerank. & -3.23 & -3.51 & -3.64 & -3.91 \\
    \rowcolor{blue!10} \quad \textit{w/o} Pre-Align. & -1.72 & -1.76 & -2.21 & -2.45 \\
    \rowcolor{blue!10} \quad \textit{w/o} Pre-Align.+ PA-Rerank. & -4.12 & -4.21 & -4.66 & -4.50 \\
    \rowcolor{blue!10} \quad \textit{w/o} Query Aug. & -2.13 & -2.31 & -2.62 & -2.87 \\
    \bottomrule
    \end{tabular}}
    \label{tab:ablation_main}
    \vspace{-0.2cm}
\end{wraptable}

\paragraph{Ablation Study.} To explore the roles of different modules in DPA-RAG. We perform an ablation study and Table~\ref{tab:ablation_main} shows the results. We use \textit{w/o} to indicate the version \textit{without} a particular module. We can see: (1) The performance of DPA-RAG declines when any component is removed, which suggests that all the components are very effective. (2) Removing the preference aligned reranker (PA-Rerank.) leads to the largest performance drop, indicating a clear knowledge preference gap between RAG components and LLMs. This confirms the beneficial of using a preference-aligned reranker for external alignment. (3) The combined performance gains of preference aligned reranker and pre-aligned task are lower than the complete DPA-RAG framework, which implies that integrating both alignment methods  yields a mutually reinforcing effect, demonstrating the superiority of our dual alignment strategies. More detailed results can be found in Appendix \ref{app:ablation}.

\subsection{Quantitative Analysis}

\begin{figure}[!t]
\centering
\begin{subfigure}[t]{0.5\linewidth}
\centering
\includegraphics[width=2.7in]{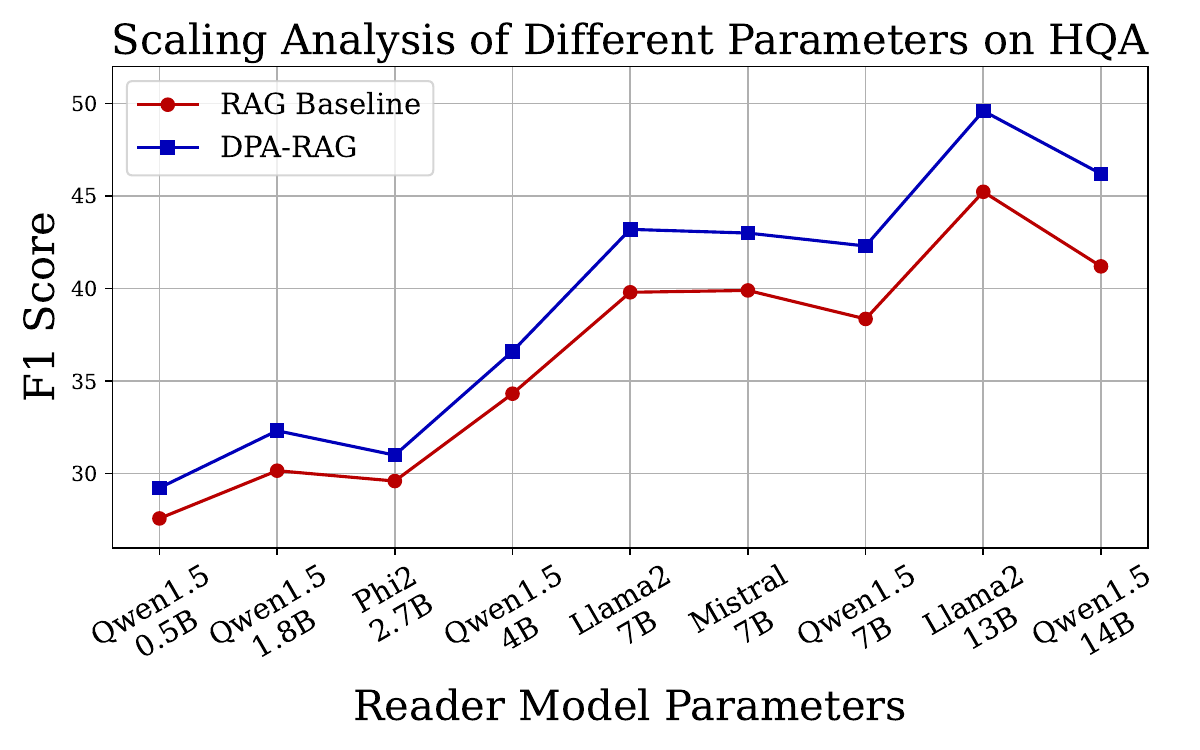}
\end{subfigure}%
\begin{subfigure}[t]{0.5\linewidth}
\centering
\includegraphics[width=2.7in]{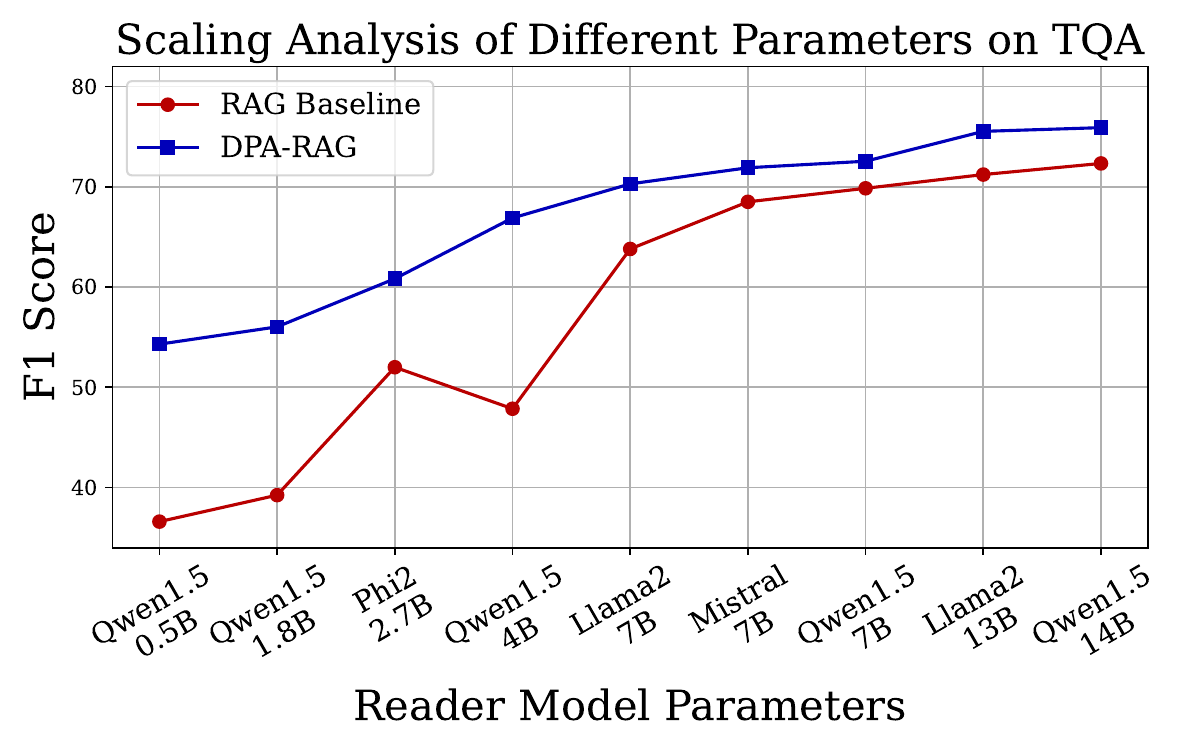}
\end{subfigure}%
\vspace{-0.3cm}
\caption{The scaling analysis of different parameter scales for HQA (left) and TQA (right).}
\label{fig:scaling}
\end{figure}

\paragraph{Scaling Analysis for Different Model Parameters.} 
To investigate the impact of parameter scale and RAG performance, we gradually increase the parameters of LLM readers (ranging from 500M to 13B) and evaluate their performance. According to the results in Figure~\ref{fig:scaling}, we have following observations:
\begin{itemize}[leftmargin=1em]
\item \textbf{Emergence of RAG Capabilities at Lower Parameter Scales (<7B)}: We notice a significant improvement in RAG baseline performance, which sharply rises from 500M to 7B parameters (40\% F1 score increase), then stabilizes for parameters beyond 7B. A similar pattern is observed in HQA, indicating a strong correlation between the emergence of RAG capabilities and model parameters. This finding presents an interesting parallel to those reported in LIMA~\citep{zhou2024lima}, where parameter increases below a certain threshold significantly boost model capabilities.

\item \textbf{Stable Performance Gains with DPA-RAG as Parameters Increase}: Compared to the baseline, DPA-RAG delivers stable improvements as parameter size expands across both datasets, displaying a smoother performance curve.

\item \textbf{Greater Benefits from DPA-RAG in Datasets with More Unalignment}: The performance gains from DPA-RAG exhibit interesting variations between TQA and HQA as parameters increase. In TQA, where the average F1 score is already over 60, the model quickly reaches a high-performance threshold as parameters increase, leaving limited room for further improvements through preference alignment. Conversely, HQA, characterized by more extensive unaligned knowledge and a lower average F1 score (below 50), shows that the alignment gains provided by DPA-RAG exceed those from increasing foundational RAG capabilities alone, leading to more improvement in alignment for RAG.
\end{itemize}

\begin{figure}[t]
\centering
\begin{subfigure}[t]{0.48\linewidth}
\centering
\includegraphics[width=2.2in]{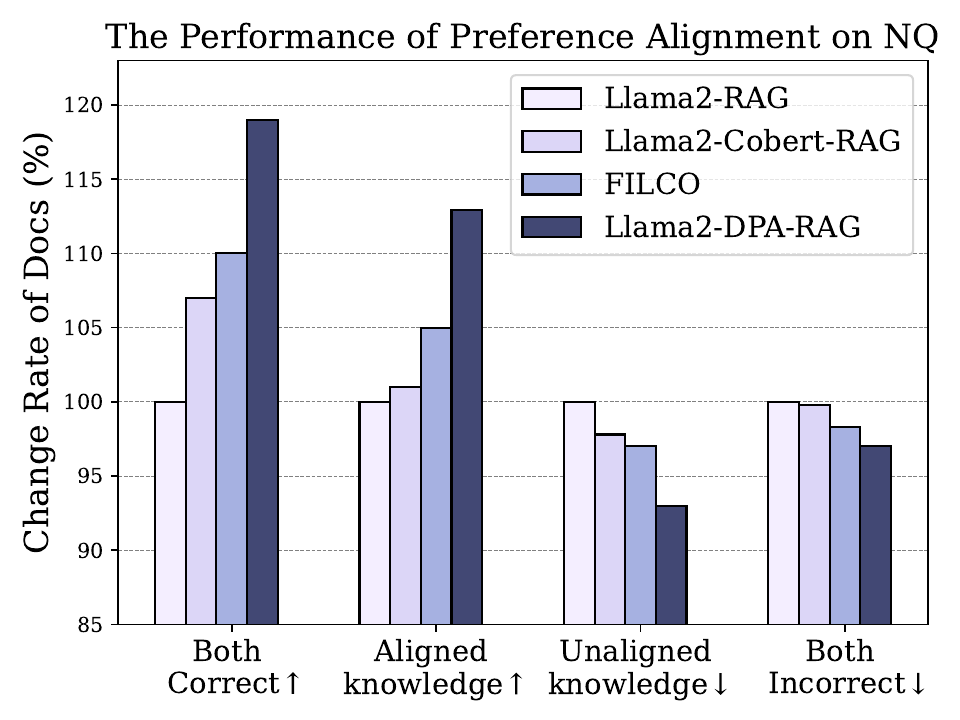}
\end{subfigure}%
\begin{subfigure}[t]{0.48\linewidth}
\centering
\includegraphics[width=2.2in]{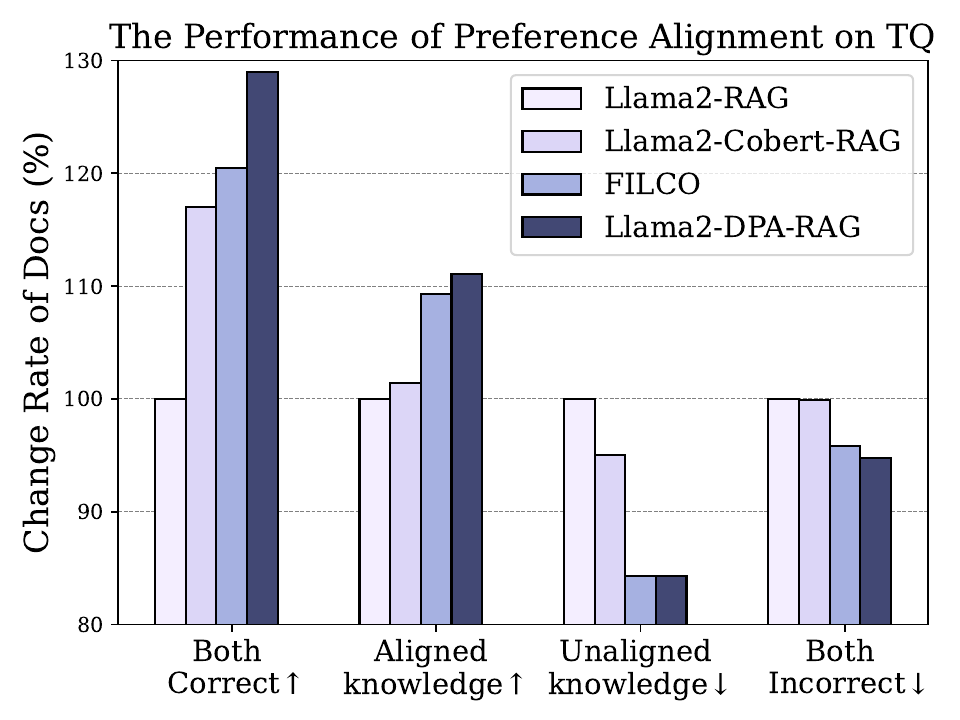}
\end{subfigure}%

\caption{The comparision experiment of preference alignment on NQ, TQA.}
\label{fig:preference}
\end{figure}


\paragraph{Effectiveness on Preference Alignment.} To delve deeper into the impact of preference alignment, in line with the setup in Section~\ref{sec:32}, we conduct a comparative experiment on direct query answering versus referencing top-3 documents. As shown in Figure~\ref{fig:preference}, DPA-RAG consistently achieves the highest scores in the category \emph{``Aligned Knowledge''} in all datasets, while significantly reducing the category \emph{``Unaligned Knowledge''}. This demonstrates that DPA-RAG effectively aligns retrieved knowledge with the LLM's inherent preferences.
Interestingly, the improvement of DPA-RAG in the ``Both Correct'' category even outperforms that observed in \emph{``Aligned Knowledge''}. Given the significant decrease in \emph{``Unaligned Knowledge''}, this suggests that DPA-RAG prioritizes addressing the conflicts present in retrieved documents. This behavior is in line with our pipeline's core principle: the preference-aligned reranker first externally eliminates misaligned knowledge, and the subsequent self-alignment stage allows the LLM to more effectively and implicitly capture information that is aligned with its preferences.

\begin{wraptable}{r}{0.45\textwidth}
 \vspace{-1.5em}
    \centering
    \scriptsize
    \caption{The performance result correlates with complexity and diversity on NQ}
        \renewcommand{\arraystretch}{1.2} 
    \setlength{\tabcolsep}{1.5mm} 
    \begin{tabular}{l|ccc|c}
    \toprule
    \textbf{Aug-Type} & \textbf{Complexity} & \textbf{Diversity} & \textbf{Total} & \textbf{NQ} \\
    \midrule
    Origin & 1.61 & 0.35 & \cellcolor{blue!6}1.96 & \cellcolor{blue!6}51.78 \\
    Rephras. & 1.64 & 0.39 & \cellcolor{blue!12}2.03 & \cellcolor{blue!12}52.27 \\
    SPARQL & 1.77 & 0.39 & \cellcolor{blue!18}2.16 & \cellcolor{blue!18}52.95 \\
    Constraint & 1.72 & 0.47 & \cellcolor{blue!24}2.19 & \cellcolor{blue!24}53.75 \\
    Decompos. & 1.77 & 0.51 & \cellcolor{blue!30}2.28 & \cellcolor{blue!30}54.16 \\
    Complexity & 1.85 & 0.48 & \cellcolor{blue!36}2.33 & \cellcolor{blue!36}54.81 \\
    \bottomrule
    \end{tabular}
    \label{tab:aug}
\vspace{-0.4cm} 
\end{wraptable}
\paragraph{Discussion on Query Augmentations.} Liu~\citep{liu2024makes} and Lu~\citep{lu2023instag} highlight the significant impact of dataset complexity and diversity on model alignment. To investigate how the complexity and diversity of our augmented queries affect RAG performance, we randomly select 1,000 samples from each dataset and employ Intag technology~\citep{lu2023instag} for automated intent annotation. For each dataset, we measure diversity by calculating $\frac{\rm\ number\ of\ unique\ tags}{\rm\ number\ of\ all\ samples}$ and complexity by $\frac{\rm\ number\ of\ all\ tags}{\rm\ number\ of\ all\ samples}$. 

\begin{figure}[t]
\centering
\begin{subfigure}[t]{0.48\linewidth}
\centering
\includegraphics[width=2.5in]{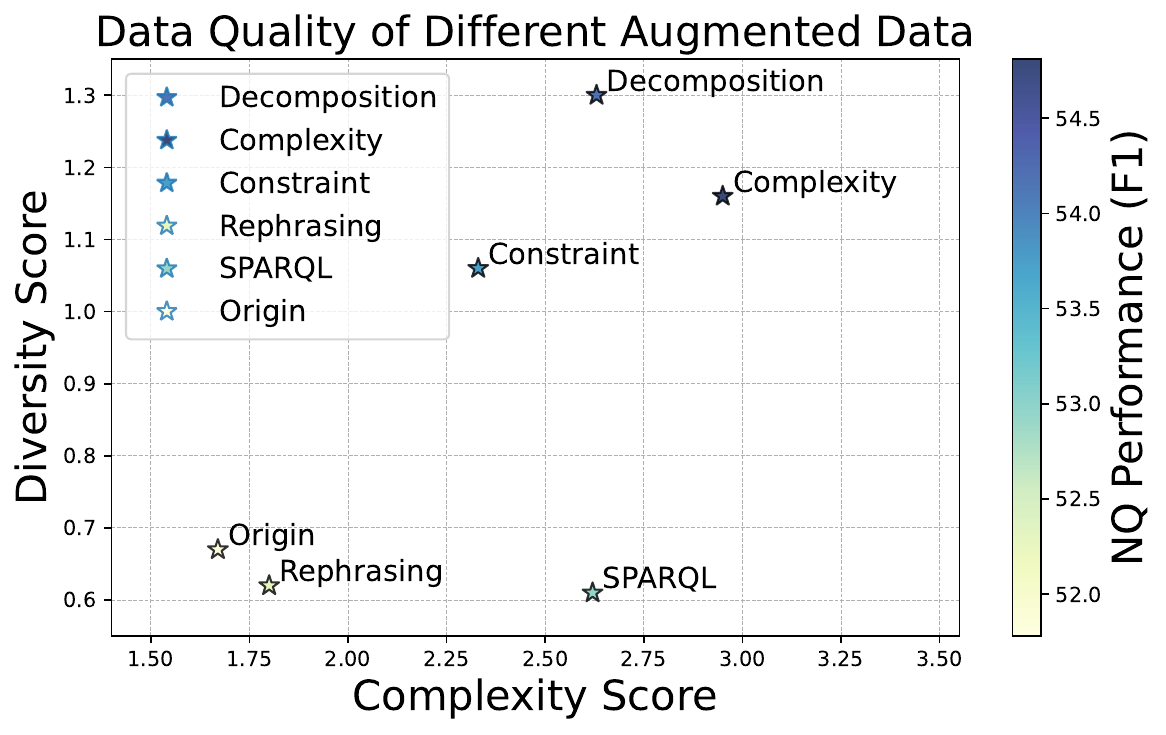}
\end{subfigure}%
\begin{subfigure}[t]{0.48\linewidth}
\centering
\includegraphics[width=2.5in]{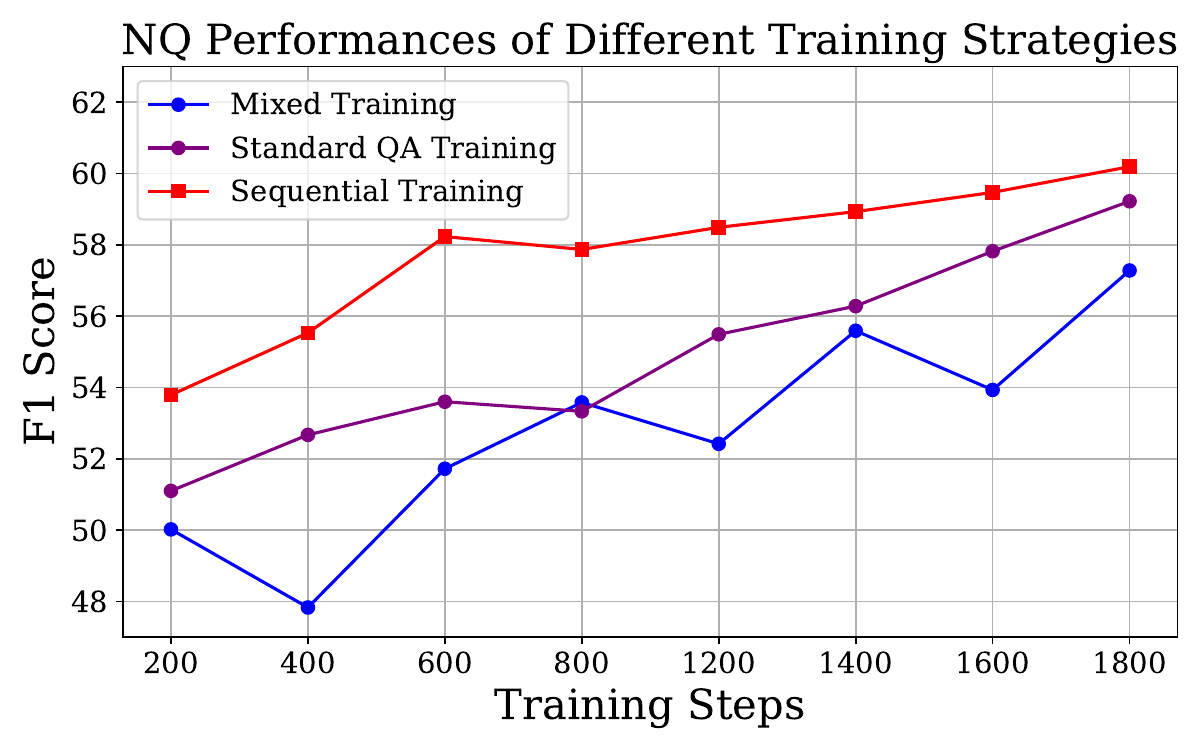}
\end{subfigure}%

\caption{The left figure illustrates the visualization of different data complexity and diversity on NQ. The right figure shows performance of different training strategies on NQ.}
\label{fig:training_step}
\end{figure}

Figure~\ref{fig:training_step} visualizes the quality of the augmented data, showing that our five methods consistently enhance data complexity. Specifically, \textit{Complexity} and \textit{Decomposition} markedly boost both complexity and diversity scores, which also align with the case studies presented in Table~\ref{app:case-study}. Moreover, we mix the augmented data with the original training set in actual proportions and calculate the data quality. 

Table~\ref{tab:aug} shows that all five augmentation strategies enhance the LLM's performance to different degrees. Surprisingly, when we sum up the two metrics, the overall trend of performance on NQ increases along with the growth of the total quality score. This insight further validates that in RAG tasks, the effectiveness of query augmentations is highly correlated with their complexity and diversity.
  

\paragraph{Sequential Training vs. Mixed Training.} In Section~\ref{sec:34}, we design a knowledge self-alignment task during the pre-aligned phase and further perform sequential SFT on the QA dataset. An alternative approach is directly mixing preference data with QA task data for joint training. Figure~\ref{fig:training_step} illustrates the performance of these two training strategies across training steps. Compared to standard QA fine-tuning, we notice that mixing training data from both tasks leads to a noticeable performance decline and fluctuations. This result may stem from optimization conflicts in multi-task training~\citep{dong2023abilities}. However, the sequential training after the pre-aligned phase yields stable performance gains, validating its efficacy. Similar conclusions have been reported in studies on reasoning~\citep{wu2024LLaMA,wang2024dolphcoder,dou2023art,tang2024mathscale}.


\section{Conclusion}

In this paper, we reveal the inherent preference gap among RAG components and first propose DPA-RAG to align diverse knowledge preferences. Specifically, we gradually extract and filter out the LLM preferred knowledge from training set, and propose five high-quality query augmentation strategies to alleviate data sparsity issues. Based on preference data, we jointly integrate pair-wise, point-wise, and contrastive preference alignment abilities into the reranker, achieving external preference alignment among RAG components. Futher, we introduce LLM Self-Alignment task to remove knowledge biases and achieve internal alignment. Experimental results demonstrate that DPA-RAG outperforms all strong baselines across four knowledge-intensive QA datasets. The extensive analysis also provides practical insights for developing reliable RAG systems.

\bibliographystyle{unsrt}
\bibliography{natbib}

\begin{thebibliography}{100}

\bibitem{ouyang2022training}
Long Ouyang, Jeffrey Wu, Xu~Jiang, Diogo Almeida, Carroll~L. Wainwright, Pamela Mishkin, Chong Zhang, Sandhini Agarwal, Katarina Slama, Alex Ray, John Schulman, Jacob Hilton, Fraser Kelton, Luke Miller, Maddie Simens, Amanda Askell, Peter Welinder, Paul~F. Christiano, Jan Leike, and Ryan Lowe.
\newblock Training language models to follow instructions with human feedback.
\newblock In Sanmi Koyejo, S.~Mohamed, A.~Agarwal, Danielle Belgrave, K.~Cho, and A.~Oh, editors, {\em Advances in Neural Information Processing Systems 35: Annual Conference on Neural Information Processing Systems 2022, NeurIPS 2022, New Orleans, LA, USA, November 28 - December 9, 2022}, 2022.

\bibitem{anil2023palm}
Rohan Anil, Andrew~M. Dai, Orhan Firat, Melvin Johnson, Dmitry Lepikhin, Alexandre Passos, Siamak Shakeri, Emanuel Taropa, Paige Bailey, Zhifeng Chen, Eric Chu, Jonathan~H. Clark, Laurent~El Shafey, Yanping Huang, Kathy Meier{-}Hellstern, Gaurav Mishra, Erica Moreira, Mark Omernick, Kevin Robinson, Sebastian Ruder, Yi~Tay, Kefan Xiao, Yuanzhong Xu, Yujing Zhang, Gustavo~Hern{\'{a}}ndez {\'{A}}brego, Junwhan Ahn, Jacob Austin, Paul Barham, Jan~A. Botha, James Bradbury, Siddhartha Brahma, Kevin Brooks, Michele Catasta, Yong Cheng, Colin Cherry, Christopher~A. Choquette{-}Choo, Aakanksha Chowdhery, Cl{\'{e}}ment Crepy, Shachi Dave, Mostafa Dehghani, Sunipa Dev, Jacob Devlin, Mark D{\'{\i}}az, Nan Du, Ethan Dyer, Vladimir Feinberg, Fangxiaoyu Feng, Vlad Fienber, Markus Freitag, Xavier Garcia, Sebastian Gehrmann, Lucas Gonzalez, and et~al.
\newblock Palm 2 technical report.
\newblock {\em CoRR}, abs/2305.10403, 2023.

\bibitem{openai2024gpt4}
OpenAI.
\newblock {GPT-4} technical report.
\newblock {\em CoRR}, abs/2303.08774, 2023.

\bibitem{chen2021evaluating}
Mark Chen, Jerry Tworek, Heewoo Jun, Qiming Yuan, Henrique~Pond{\'{e}} de~Oliveira~Pinto, Jared Kaplan, Harrison Edwards, Yuri Burda, Nicholas Joseph, Greg Brockman, Alex Ray, Raul Puri, Gretchen Krueger, Michael Petrov, Heidy Khlaaf, Girish Sastry, Pamela Mishkin, Brooke Chan, Scott Gray, Nick Ryder, Mikhail Pavlov, Alethea Power, Lukasz Kaiser, Mohammad Bavarian, Clemens Winter, Philippe Tillet, Felipe~Petroski Such, Dave Cummings, Matthias Plappert, Fotios Chantzis, Elizabeth Barnes, Ariel Herbert{-}Voss, William~Hebgen Guss, Alex Nichol, Alex Paino, Nikolas Tezak, Jie Tang, Igor Babuschkin, Suchir Balaji, Shantanu Jain, William Saunders, Christopher Hesse, Andrew~N. Carr, Jan Leike, Joshua Achiam, Vedant Misra, Evan Morikawa, Alec Radford, Matthew Knight, Miles Brundage, Mira Murati, Katie Mayer, Peter Welinder, Bob McGrew, Dario Amodei, Sam McCandlish, Ilya Sutskever, and Wojciech Zaremba.
\newblock Evaluating large language models trained on code.
\newblock {\em CoRR}, abs/2107.03374, 2021.

\bibitem{longpre2023flan}
Shayne Longpre, Le~Hou, Tu~Vu, Albert Webson, Hyung~Won Chung, Yi~Tay, Denny Zhou, Quoc~V. Le, Barret Zoph, Jason Wei, and Adam Roberts.
\newblock The flan collection: Designing data and methods for effective instruction tuning.
\newblock In Andreas Krause, Emma Brunskill, Kyunghyun Cho, Barbara Engelhardt, Sivan Sabato, and Jonathan Scarlett, editors, {\em International Conference on Machine Learning, {ICML} 2023, 23-29 July 2023, Honolulu, Hawaii, {USA}}, volume 202 of {\em Proceedings of Machine Learning Research}, pages 22631--22648. {PMLR}, 2023.

\bibitem{wei2023chainofthought}
Jason Wei, Xuezhi Wang, Dale Schuurmans, Maarten Bosma, Brian Ichter, Fei Xia, Ed~H. Chi, Quoc~V. Le, and Denny Zhou.
\newblock Chain-of-thought prompting elicits reasoning in large language models.
\newblock In Sanmi Koyejo, S.~Mohamed, A.~Agarwal, Danielle Belgrave, K.~Cho, and A.~Oh, editors, {\em Advances in Neural Information Processing Systems 35: Annual Conference on Neural Information Processing Systems 2022, NeurIPS 2022, New Orleans, LA, USA, November 28 - December 9, 2022}, 2022.

\bibitem{luo2023wizardmath}
Haipeng Luo, Qingfeng Sun, Can Xu, Pu~Zhao, Jianguang Lou, Chongyang Tao, Xiubo Geng, Qingwei Lin, Shifeng Chen, and Dongmei Zhang.
\newblock Wizardmath: Empowering mathematical reasoning for large language models via reinforced evol-instruct.
\newblock {\em CoRR}, abs/2308.09583, 2023.

\bibitem{luo2023wizardcoder}
Ziyang Luo, Can Xu, Pu~Zhao, Qingfeng Sun, Xiubo Geng, Wenxiang Hu, Chongyang Tao, Jing Ma, Qingwei Lin, and Daxin Jiang.
\newblock Wizardcoder: Empowering code large language models with evol-instruct.
\newblock {\em CoRR}, abs/2306.08568, 2023.

\bibitem{yuan2023scaling}
Zheng Yuan, Hongyi Yuan, Chengpeng Li, Guanting Dong, Chuanqi Tan, and Chang Zhou.
\newblock Scaling relationship on learning mathematical reasoning with large language models.
\newblock {\em arXiv preprint arXiv:2308.01825}, 2023.

\bibitem{bang2023multitask}
Yejin Bang, Samuel Cahyawijaya, Nayeon Lee, Wenliang Dai, Dan Su, Bryan Wilie, Holy Lovenia, Ziwei Ji, Tiezheng Yu, Willy Chung, Quyet~V. Do, Yan Xu, and Pascale Fung.
\newblock A multitask, multilingual, multimodal evaluation of chatgpt on reasoning, hallucination, and interactivity.
\newblock In Jong~C. Park, Yuki Arase, Baotian Hu, Wei Lu, Derry Wijaya, Ayu Purwarianti, and Adila~Alfa Krisnadhi, editors, {\em Proceedings of the 13th International Joint Conference on Natural Language Processing and the 3rd Conference of the Asia-Pacific Chapter of the Association for Computational Linguistics, {IJCNLP} 2023 -Volume 1: Long Papers, Nusa Dua, Bali, November 1 - 4, 2023}, pages 675--718. Association for Computational Linguistics, 2023.

\bibitem{zhang2023sirens}
Yue Zhang, Yafu Li, Leyang Cui, Deng Cai, Lemao Liu, Tingchen Fu, Xinting Huang, Enbo Zhao, Yu~Zhang, Yulong Chen, Longyue Wang, Anh~Tuan Luu, Wei Bi, Freda Shi, and Shuming Shi.
\newblock Siren's song in the {AI} ocean: {A} survey on hallucination in large language models.
\newblock {\em CoRR}, abs/2309.01219, 2023.

\bibitem{guu2020realm}
Kelvin Guu, Kenton Lee, Zora Tung, Panupong Pasupat, and Ming{-}Wei Chang.
\newblock {REALM:} retrieval-augmented language model pre-training.
\newblock {\em CoRR}, abs/2002.08909, 2020.

\bibitem{lewis2021retrievalaugmented}
Patrick S.~H. Lewis, Ethan Perez, Aleksandra Piktus, Fabio Petroni, Vladimir Karpukhin, Naman Goyal, Heinrich K{\"{u}}ttler, Mike Lewis, Wen{-}tau Yih, Tim Rockt{\"{a}}schel, Sebastian Riedel, and Douwe Kiela.
\newblock Retrieval-augmented generation for knowledge-intensive {NLP} tasks.
\newblock In Hugo Larochelle, Marc'Aurelio Ranzato, Raia Hadsell, Maria{-}Florina Balcan, and Hsuan{-}Tien Lin, editors, {\em Advances in Neural Information Processing Systems 33: Annual Conference on Neural Information Processing Systems 2020, NeurIPS 2020, December 6-12, 2020, virtual}, 2020.

\bibitem{press-etal-2023-measuring}
Ofir Press, Muru Zhang, Sewon Min, Ludwig Schmidt, Noah~A. Smith, and Mike Lewis.
\newblock Measuring and narrowing the compositionality gap in language models.
\newblock In Houda Bouamor, Juan Pino, and Kalika Bali, editors, {\em Findings of the Association for Computational Linguistics: {EMNLP} 2023, Singapore, December 6-10, 2023}, pages 5687--5711. Association for Computational Linguistics, 2023.

\bibitem{bai2023griprank}
Jiaqi Bai, Hongcheng Guo, Jiaheng Liu, Jian Yang, Xinnian Liang, Zhao Yan, and Zhoujun Li.
\newblock Griprank: Bridging the gap between retrieval and generation via the generative knowledge improved passage ranking.
\newblock In Ingo Frommholz, Frank Hopfgartner, Mark Lee, Michael Oakes, Mounia Lalmas, Min Zhang, and Rodrygo L.~T. Santos, editors, {\em Proceedings of the 32nd {ACM} International Conference on Information and Knowledge Management, {CIKM} 2023, Birmingham, United Kingdom, October 21-25, 2023}, pages 36--46. {ACM}, 2023.

\bibitem{li2022survey}
Huayang Li, Yixuan Su, Deng Cai, Yan Wang, and Lemao Liu.
\newblock A survey on retrieval-augmented text generation.
\newblock {\em CoRR}, abs/2202.01110, 2022.

\bibitem{petroni2019language}
Fabio Petroni, Tim Rockt{\"{a}}schel, Sebastian Riedel, Patrick S.~H. Lewis, Anton Bakhtin, Yuxiang Wu, and Alexander~H. Miller.
\newblock Language models as knowledge bases?
\newblock In Kentaro Inui, Jing Jiang, Vincent Ng, and Xiaojun Wan, editors, {\em Proceedings of the 2019 Conference on Empirical Methods in Natural Language Processing and the 9th International Joint Conference on Natural Language Processing, {EMNLP-IJCNLP} 2019, Hong Kong, China, November 3-7, 2019}, pages 2463--2473. Association for Computational Linguistics, 2019.

\bibitem{pmlr-v133-min21a}
Sewon Min, Jordan~L. Boyd{-}Graber, Chris Alberti, Danqi Chen, Eunsol Choi, Michael Collins, Kelvin Guu, Hannaneh Hajishirzi, Kenton Lee, Jennimaria Palomaki, Colin Raffel, Adam Roberts, Tom Kwiatkowski, Patrick S.~H. Lewis, Yuxiang Wu, Heinrich K{\"{u}}ttler, Linqing Liu, Pasquale Minervini, Pontus Stenetorp, Sebastian Riedel, Sohee Yang, Minjoon Seo, Gautier Izacard, Fabio Petroni, Lucas Hosseini, Nicola~De Cao, Edouard Grave, Ikuya Yamada, Sonse Shimaoka, Masatoshi Suzuki, Shumpei Miyawaki, Shun Sato, Ryo Takahashi, Jun Suzuki, Martin Fajcik, Martin Docekal, Karel Ondrej, Pavel Smrz, Hao Cheng, Yelong Shen, Xiaodong Liu, Pengcheng He, Weizhu Chen, Jianfeng Gao, Barlas Oguz, Xilun Chen, Vladimir Karpukhin, Stan Peshterliev, Dmytro Okhonko, Michael~Sejr Schlichtkrull, Sonal Gupta, Yashar Mehdad, and Wen{-}tau Yih.
\newblock Neurips 2020 efficientqa competition: Systems, analyses and lessons learned.
\newblock In Hugo~Jair Escalante and Katja Hofmann, editors, {\em NeurIPS 2020 Competition and Demonstration Track, 6-12 December 2020, Virtual Event / Vancouver, BC, Canada}, volume 133 of {\em Proceedings of Machine Learning Research}, pages 86--111. {PMLR}, 2020.

\bibitem{lebovitz2021ai}
Sarah Lebovitz, Natalia Levina, and Hila Lifshitz{-}Assaf.
\newblock Is {AI} ground truth really true? the dangers of training and evaluating {AI} tools based on experts' know-what.
\newblock {\em {MIS} Q.}, 45(3), 2021.

\bibitem{sener2018multi}
Ozan Sener and Vladlen Koltun.
\newblock Multi-task learning as multi-objective optimization.
\newblock In Samy Bengio, Hanna~M. Wallach, Hugo Larochelle, Kristen Grauman, Nicol{\`{o}} Cesa{-}Bianchi, and Roman Garnett, editors, {\em Advances in Neural Information Processing Systems 31: Annual Conference on Neural Information Processing Systems 2018, NeurIPS 2018, December 3-8, 2018, Montr{\'{e}}al, Canada}, pages 525--536, 2018.

\bibitem{ji2024ai}
Jiaming Ji, Tianyi Qiu, Boyuan Chen, Borong Zhang, Hantao Lou, Kaile Wang, Yawen Duan, Zhonghao He, Jiayi Zhou, Zhaowei Zhang, Fanzhi Zeng, Kwan~Yee Ng, Juntao Dai, Xuehai Pan, Aidan O'Gara, Yingshan Lei, Hua Xu, Brian Tse, Jie Fu, Stephen McAleer, Yaodong Yang, Yizhou Wang, Song{-}Chun Zhu, Yike Guo, and Wen Gao.
\newblock {AI} alignment: {A} comprehensive survey.
\newblock {\em CoRR}, abs/2310.19852, 2023.

\bibitem{fang2024domainagnostic}
Yin Fang, Ningyu Zhang, Zhuo Chen, Lingbing Guo, Xiaohui Fan, and Huajun Chen.
\newblock Domain-agnostic molecular generation with chemical feedback, 2024.

\bibitem{wang2023aligning}
Yufei Wang, Wanjun Zhong, Liangyou Li, Fei Mi, Xingshan Zeng, Wenyong Huang, Lifeng Shang, Xin Jiang, and Qun Liu.
\newblock Aligning large language models with human: {A} survey.
\newblock {\em CoRR}, abs/2307.12966, 2023.

\bibitem{jiang-etal-2023-structgpt}
Jinhao Jiang, Kun Zhou, Zican Dong, Keming Ye, Xin Zhao, and Ji{-}Rong Wen.
\newblock Structgpt: {A} general framework for large language model to reason over structured data.
\newblock In Houda Bouamor, Juan Pino, and Kalika Bali, editors, {\em Proceedings of the 2023 Conference on Empirical Methods in Natural Language Processing, {EMNLP} 2023, Singapore, December 6-10, 2023}, pages 9237--9251. Association for Computational Linguistics, 2023.

\bibitem{DBLP:journals/corr/ppo}
John Schulman, Filip Wolski, Prafulla Dhariwal, Alec Radford, and Oleg Klimov.
\newblock Proximal policy optimization algorithms.
\newblock {\em CoRR}, abs/1707.06347, 2017.

\bibitem{liu2023alignment-add6}
Hao Liu, Carmelo Sferrazza, and Pieter Abbeel.
\newblock Chain of hindsight aligns language models with feedback.
\newblock {\em CoRR}, abs/2302.02676, 2023.

\bibitem{DBLP:journals/corr/alignment-add2}
Ruibo Liu, Ruixin Yang, Chenyan Jia, Ge~Zhang, Denny Zhou, Andrew~M. Dai, Diyi Yang, and Soroush Vosoughi.
\newblock Training socially aligned language models in simulated human society.
\newblock {\em CoRR}, abs/2305.16960, 2023.

\bibitem{DBLP:journals/corr/alignment-add4}
Tianqi Liu, Yao Zhao, Rishabh Joshi, Misha Khalman, Mohammad Saleh, Peter~J. Liu, and Jialu Liu.
\newblock Statistical rejection sampling improves preference optimization.
\newblock {\em CoRR}, abs/2309.06657, 2023.

\bibitem{nathani2023alignment}
Deepak Nathani, David Wang, Liangming Pan, and William~Yang Wang.
\newblock {MAF:} multi-aspect feedback for improving reasoning in large language models.
\newblock In Houda Bouamor, Juan Pino, and Kalika Bali, editors, {\em Proceedings of the 2023 Conference on Empirical Methods in Natural Language Processing, {EMNLP} 2023, Singapore, December 6-10, 2023}, pages 6591--6616. Association for Computational Linguistics, 2023.

\bibitem{DBLP:journals/corr/alignment-add1}
Bo~Shen, Jiaxin Zhang, Taihong Chen, Daoguang Zan, Bing Geng, An~Fu, Muhan Zeng, Ailun Yu, Jichuan Ji, Jingyang Zhao, Yuenan Guo, and Qianxiang Wang.
\newblock Pangu-coder2: Boosting large language models for code with ranking feedback.
\newblock {\em CoRR}, abs/2307.14936, 2023.

\bibitem{DBLP:journals/corr/alignment-add5}
Zeqiu Wu, Yushi Hu, Weijia Shi, Nouha Dziri, Alane Suhr, Prithviraj Ammanabrolu, Noah~A. Smith, Mari Ostendorf, and Hannaneh Hajishirzi.
\newblock Fine-grained human feedback gives better rewards for language model training.
\newblock {\em CoRR}, abs/2306.01693, 2023.

\bibitem{DBLP:journals/corr/alignment-add3}
Weizhe Yuan, Kyunghyun Cho, and Jason Weston.
\newblock System-level natural language feedback.
\newblock {\em CoRR}, abs/2306.13588, 2023.

\bibitem{DBLP:journals/corr/slic-hf}
Yao Zhao, Rishabh Joshi, Tianqi Liu, Misha Khalman, Mohammad Saleh, and Peter~J. Liu.
\newblock Slic-hf: Sequence likelihood calibration with human feedback.
\newblock {\em CoRR}, abs/2305.10425, 2023.

\bibitem{song2024preference}
Feifan Song, Bowen Yu, Minghao Li, Haiyang Yu, Fei Huang, Yongbin Li, and Houfeng Wang.
\newblock Preference ranking optimization for human alignment.
\newblock In Michael~J. Wooldridge, Jennifer~G. Dy, and Sriraam Natarajan, editors, {\em Thirty-Eighth {AAAI} Conference on Artificial Intelligence, {AAAI} 2024, Thirty-Sixth Conference on Innovative Applications of Artificial Intelligence, {IAAI} 2024, Fourteenth Symposium on Educational Advances in Artificial Intelligence, {EAAI} 2014, February 20-27, 2024, Vancouver, Canada}, pages 18990--18998. {AAAI} Press, 2024.

\bibitem{rafailov2023direct}
Rafael Rafailov, Archit Sharma, Eric Mitchell, Christopher~D. Manning, Stefano Ermon, and Chelsea Finn.
\newblock Direct preference optimization: Your language model is secretly a reward model.
\newblock In Alice Oh, Tristan Naumann, Amir Globerson, Kate Saenko, Moritz Hardt, and Sergey Levine, editors, {\em Advances in Neural Information Processing Systems 36: Annual Conference on Neural Information Processing Systems 2023, NeurIPS 2023, New Orleans, LA, USA, December 10 - 16, 2023}, 2023.

\bibitem{shi2023replug}
Weijia Shi, Sewon Min, Michihiro Yasunaga, Minjoon Seo, Rich James, Mike Lewis, Luke Zettlemoyer, and Wen{-}tau Yih.
\newblock {REPLUG:} retrieval-augmented black-box language models.
\newblock {\em CoRR}, abs/2301.12652, 2023.

\bibitem{bonifacio2022inpars}
Luiz~Henrique Bonifacio, Hugo~Queiroz Abonizio, Marzieh Fadaee, and Rodrigo~Frassetto Nogueira.
\newblock Inpars: Data augmentation for information retrieval using large language models.
\newblock {\em CoRR}, abs/2202.05144, 2022.

\bibitem{jeronymo2023inparsv2}
Vitor Jeronymo, Luiz~Henrique Bonifacio, Hugo~Queiroz Abonizio, Marzieh Fadaee, Roberto de~Alencar~Lotufo, Jakub Zavrel, and Rodrigo~Frassetto Nogueira.
\newblock Inpars-v2: Large language models as efficient dataset generators for information retrieval.
\newblock {\em CoRR}, abs/2301.01820, 2023.

\bibitem{jiang2023active}
Zhengbao Jiang, Frank~F. Xu, Luyu Gao, Zhiqing Sun, Qian Liu, Jane Dwivedi{-}Yu, Yiming Yang, Jamie Callan, and Graham Neubig.
\newblock Active retrieval augmented generation.
\newblock In Houda Bouamor, Juan Pino, and Kalika Bali, editors, {\em Proceedings of the 2023 Conference on Empirical Methods in Natural Language Processing, {EMNLP} 2023, Singapore, December 6-10, 2023}, pages 7969--7992. Association for Computational Linguistics, 2023.

\bibitem{yao2023react}
Shunyu Yao, Jeffrey Zhao, Dian Yu, Nan Du, Izhak Shafran, Karthik~R. Narasimhan, and Yuan Cao.
\newblock React: Synergizing reasoning and acting in language models.
\newblock In {\em The Eleventh International Conference on Learning Representations, {ICLR} 2023, Kigali, Rwanda, May 1-5, 2023}. OpenReview.net, 2023.

\bibitem{ren2023investigating}
Ruiyang Ren, Yuhao Wang, Yingqi Qu, Wayne~Xin Zhao, Jing Liu, Hao Tian, Hua Wu, Ji{-}Rong Wen, and Haifeng Wang.
\newblock Investigating the factual knowledge boundary of large language models with retrieval augmentation.
\newblock {\em CoRR}, abs/2307.11019, 2023.

\bibitem{zhou2024metacognitive}
Yujia Zhou, Zheng Liu, Jiajie Jin, Jian{-}Yun Nie, and Zhicheng Dou.
\newblock Metacognitive retrieval-augmented large language models.
\newblock {\em CoRR}, abs/2402.11626, 2024.

\bibitem{trivedi2023interleaving}
Harsh Trivedi, Niranjan Balasubramanian, Tushar Khot, and Ashish Sabharwal.
\newblock Interleaving retrieval with chain-of-thought reasoning for knowledge-intensive multi-step questions.
\newblock In Anna Rogers, Jordan~L. Boyd{-}Graber, and Naoaki Okazaki, editors, {\em Proceedings of the 61st Annual Meeting of the Association for Computational Linguistics (Volume 1: Long Papers), {ACL} 2023, Toronto, Canada, July 9-14, 2023}, pages 10014--10037. Association for Computational Linguistics, 2023.

\bibitem{wang2024llms}
Keheng Wang, Feiyu Duan, Peiguang Li, Sirui Wang, and Xunliang Cai.
\newblock Llms know what they need: Leveraging a missing information guided framework to empower retrieval-augmented generation, 2024.

\bibitem{zhang2024raft}
Tianjun Zhang, Shishir~G. Patil, Naman Jain, Sheng Shen, Matei Zaharia, Ion Stoica, and Joseph~E. Gonzalez.
\newblock {RAFT:} adapting language model to domain specific {RAG}.
\newblock {\em CoRR}, abs/2403.10131, 2024.

\bibitem{wang2023learning}
Zhiruo Wang, Jun Araki, Zhengbao Jiang, Md.~Rizwan Parvez, and Graham Neubig.
\newblock Learning to filter context for retrieval-augmented generation.
\newblock {\em CoRR}, abs/2311.08377, 2023.

\bibitem{zhang2024knowledgeable}
Yichi Zhang, Zhuo Chen, Yin Fang, Yanxi Lu, Fangming Li, Wen Zhang, and Huajun Chen.
\newblock Knowledgeable preference alignment for llms in domain-specific question answering, 2024.

\bibitem{jin2024bider}
Jiajie Jin, Yutao Zhu, Yujia Zhou, and Zhicheng Dou.
\newblock {BIDER:} bridging knowledge inconsistency for efficient retrieval-augmented llms via key supporting evidence.
\newblock {\em CoRR}, abs/2402.12174, 2024.

\bibitem{wang2024rat}
Zihao Wang, Anji Liu, Haowei Lin, Jiaqi Li, Xiaojian Ma, and Yitao Liang.
\newblock {RAT:} retrieval augmented thoughts elicit context-aware reasoning in long-horizon generation.
\newblock {\em CoRR}, abs/2403.05313, 2024.

\bibitem{reimers2019sentencebert}
Nils Reimers and Iryna Gurevych.
\newblock Sentence-bert: Sentence embeddings using siamese bert-networks.
\newblock In Kentaro Inui, Jing Jiang, Vincent Ng, and Xiaojun Wan, editors, {\em Proceedings of the 2019 Conference on Empirical Methods in Natural Language Processing and the 9th International Joint Conference on Natural Language Processing, {EMNLP-IJCNLP} 2019, Hong Kong, China, November 3-7, 2019}, pages 3980--3990. Association for Computational Linguistics, 2019.

\bibitem{bge_embedding}
Shitao Xiao, Zheng Liu, Peitian Zhang, and Niklas Muennighof.
\newblock C-pack: Packaged resources to advance general chinese embedding.
\newblock {\em CoRR}, abs/2309.07597, 2023.

\bibitem{khattab2020colbert}
Omar Khattab and Matei Zaharia.
\newblock Colbert: Efficient and effective passage search via contextualized late interaction over {BERT}.
\newblock In Jimmy~X. Huang, Yi~Chang, Xueqi Cheng, Jaap Kamps, Vanessa Murdock, Ji{-}Rong Wen, and Yiqun Liu, editors, {\em Proceedings of the 43rd International {ACM} {SIGIR} conference on research and development in Information Retrieval, {SIGIR} 2020, Virtual Event, China, July 25-30, 2020}, pages 39--48. {ACM}, 2020.

\bibitem{nogueira2019multistage}
Rodrigo~Frassetto Nogueira, Wei Yang, Kyunghyun Cho, and Jimmy Lin.
\newblock Multi-stage document ranking with {BERT}.
\newblock {\em CoRR}, abs/1910.14424, 2019.

\bibitem{liu2023pre}
Pengfei Liu, Weizhe Yuan, Jinlan Fu, Zhengbao Jiang, Hiroaki Hayashi, and Graham Neubig.
\newblock Pre-train, prompt, and predict: {A} systematic survey of prompting methods in natural language processing.
\newblock {\em {ACM} Comput. Surv.}, 55(9):195:1--195:35, 2023.

\bibitem{nogueira-etal-2020-document}
Rodrigo~Frassetto Nogueira, Zhiying Jiang, Ronak Pradeep, and Jimmy Lin.
\newblock Document ranking with a pretrained sequence-to-sequence model.
\newblock In Trevor Cohn, Yulan He, and Yang Liu, editors, {\em Findings of the Association for Computational Linguistics: {EMNLP} 2020, Online Event, 16-20 November 2020}, volume {EMNLP} 2020 of {\em Findings of {ACL}}, pages 708--718. Association for Computational Linguistics, 2020.

\bibitem{ju2022texttotext}
Jia{-}Huei Ju, Jheng{-}Hong Yang, and Chuan{-}Ju Wang.
\newblock Text-to-text multi-view learning for passage re-ranking.
\newblock In Fernando Diaz, Chirag Shah, Torsten Suel, Pablo Castells, Rosie Jones, and Tetsuya Sakai, editors, {\em {SIGIR} '21: The 44th International {ACM} {SIGIR} Conference on Research and Development in Information Retrieval, Virtual Event, Canada, July 11-15, 2021}, pages 1803--1807. {ACM}, 2021.

\bibitem{pradeep2021expandomonoduo}
Ronak Pradeep, Rodrigo~Frassetto Nogueira, and Jimmy Lin.
\newblock The expando-mono-duo design pattern for text ranking with pretrained sequence-to-sequence models.
\newblock {\em CoRR}, abs/2101.05667, 2021.

\bibitem{zhuang2022rankt5}
Honglei Zhuang, Zhen Qin, Rolf Jagerman, Kai Hui, Ji~Ma, Jing Lu, Jianmo Ni, Xuanhui Wang, and Michael Bendersky.
\newblock Rankt5: Fine-tuning {T5} for text ranking with ranking losses.
\newblock In Hsin{-}Hsi Chen, Wei{-}Jou~(Edward) Duh, Hen{-}Hsen Huang, Makoto~P. Kato, Josiane Mothe, and Barbara Poblete, editors, {\em Proceedings of the 46th International {ACM} {SIGIR} Conference on Research and Development in Information Retrieval, {SIGIR} 2023, Taipei, Taiwan, July 23-27, 2023}, pages 2308--2313. {ACM}, 2023.

\bibitem{sun2023chatgpt}
Weiwei Sun, Lingyong Yan, Xinyu Ma, Shuaiqiang Wang, Pengjie Ren, Zhumin Chen, Dawei Yin, and Zhaochun Ren.
\newblock Is chatgpt good at search? investigating large language models as re-ranking agents.
\newblock In Houda Bouamor, Juan Pino, and Kalika Bali, editors, {\em Proceedings of the 2023 Conference on Empirical Methods in Natural Language Processing, {EMNLP} 2023, Singapore, December 6-10, 2023}, pages 14918--14937. Association for Computational Linguistics, 2023.

\bibitem{qin2024large}
Zhen Qin, Rolf Jagerman, Kai Hui, Honglei Zhuang, Junru Wu, Jiaming Shen, Tianqi Liu, Jialu Liu, Donald Metzler, Xuanhui Wang, and Michael Bendersky.
\newblock Large language models are effective text rankers with pairwise ranking prompting.
\newblock {\em CoRR}, abs/2306.17563, 2023.

\bibitem{ma2023zero}
Xueguang Ma, Xinyu Zhang, Ronak Pradeep, and Jimmy Lin.
\newblock Zero-shot listwise document reranking with a large language model.
\newblock {\em CoRR}, abs/2305.02156, 2023.

\bibitem{ma2023finetuning}
Xueguang Ma, Liang Wang, Nan Yang, Furu Wei, and Jimmy Lin.
\newblock Fine-tuning llama for multi-stage text retrieval.
\newblock {\em CoRR}, abs/2310.08319, 2023.

\bibitem{pei2019personalized}
Changhua Pei, Yi~Zhang, Yongfeng Zhang, Fei Sun, Xiao Lin, Hanxiao Sun, Jian Wu, Peng Jiang, Junfeng Ge, Wenwu Ou, and Dan Pei.
\newblock Personalized re-ranking for recommendation.
\newblock In Toine Bogers, Alan Said, Peter Brusilovsky, and Domonkos Tikk, editors, {\em Proceedings of the 13th {ACM} Conference on Recommender Systems, RecSys 2019, Copenhagen, Denmark, September 16-20, 2019}, pages 3--11. {ACM}, 2019.

\bibitem{li2022pear}
Yi~Li, Jieming Zhu, Weiwen Liu, Liangcai Su, Guohao Cai, Qi~Zhang, Ruiming Tang, Xi~Xiao, and Xiuqiang He.
\newblock {PEAR:} personalized re-ranking with contextualized transformer for recommendation.
\newblock In Fr{\'{e}}d{\'{e}}rique Laforest, Rapha{\"{e}}l Troncy, Elena Simperl, Deepak Agarwal, Aristides Gionis, Ivan Herman, and Lionel M{\'{e}}dini, editors, {\em Companion of The Web Conference 2022, Virtual Event / Lyon, France, April 25 - 29, 2022}, pages 62--66. {ACM}, 2022.

\bibitem{saadfalcon2023udapdr}
Jon Saad{-}Falcon, Omar Khattab, Keshav Santhanam, Radu Florian, Martin Franz, Salim Roukos, Avirup Sil, Md.~Arafat Sultan, and Christopher Potts.
\newblock {UDAPDR:} unsupervised domain adaptation via {LLM} prompting and distillation of rerankers.
\newblock In Houda Bouamor, Juan Pino, and Kalika Bali, editors, {\em Proceedings of the 2023 Conference on Empirical Methods in Natural Language Processing, {EMNLP} 2023, Singapore, December 6-10, 2023}, pages 11265--11279. Association for Computational Linguistics, 2023.

\bibitem{ma2023large}
Yubo Ma, Yixin Cao, Yong Hong, and Aixin Sun.
\newblock Large language model is not a good few-shot information extractor, but a good reranker for hard samples!
\newblock In Houda Bouamor, Juan Pino, and Kalika Bali, editors, {\em Findings of the Association for Computational Linguistics: {EMNLP} 2023, Singapore, December 6-10, 2023}, pages 10572--10601. Association for Computational Linguistics, 2023.

\bibitem{shi2022xricl}
Peng Shi, Rui Zhang, He~Bai, and Jimmy Lin.
\newblock {XRICL:} cross-lingual retrieval-augmented in-context learning for cross-lingual text-to-sql semantic parsing.
\newblock In Yoav Goldberg, Zornitsa Kozareva, and Yue Zhang, editors, {\em Findings of the Association for Computational Linguistics: {EMNLP} 2022, Abu Dhabi, United Arab Emirates, December 7-11, 2022}, pages 5248--5259. Association for Computational Linguistics, 2022.

\bibitem{liu2024makes}
Wei Liu, Weihao Zeng, Keqing He, Yong Jiang, and Junxian He.
\newblock What makes good data for alignment? {A} comprehensive study of automatic data selection in instruction tuning.
\newblock {\em CoRR}, abs/2312.15685, 2023.

\bibitem{zeng2024automatic}
Weihao Zeng, Can Xu, Yingxiu Zhao, Jian-Guang Lou, and Weizhu Chen.
\newblock Automatic instruction evolving for large language models, 2024.

\bibitem{yu2023metamath}
Longhui Yu, Weisen Jiang, Han Shi, Jincheng Yu, Zhengying Liu, Yu~Zhang, James~T. Kwok, Zhenguo Li, Adrian Weller, and Weiyang Liu.
\newblock Metamath: Bootstrap your own mathematical questions for large language models.
\newblock {\em CoRR}, abs/2309.12284, 2023.

\bibitem{li2023query}
Chengpeng Li, Zheng Yuan, Guanting Dong, Keming Lu, Jiancan Wu, Chuanqi Tan, Xiang Wang, and Chang Zhou.
\newblock Query and response augmentation cannot help out-of-domain math reasoning generalization.
\newblock {\em arXiv preprint arXiv:2310.05506}, 2023.

\bibitem{shannon1948mathematical}
Claude~E. Shannon.
\newblock A mathematical theory of communication.
\newblock {\em Bell Syst. Tech. J.}, 27(3):379--423, 1948.

\bibitem{zhuang2023open}
Shengyao Zhuang, Bing Liu, Bevan Koopman, and Guido Zuccon.
\newblock Open-source large language models are strong zero-shot query likelihood models for document ranking.
\newblock {\em arXiv preprint arXiv:2310.13243}, 2023.

\bibitem{stiennon2022learning}
Nisan Stiennon, Long Ouyang, Jeff Wu, Daniel~M. Ziegler, Ryan Lowe, Chelsea Voss, Alec Radford, Dario Amodei, and Paul~F. Christiano.
\newblock Learning to summarize from human feedback.
\newblock {\em CoRR}, abs/2009.01325, 2020.

\bibitem{oord2019representation}
A{\"{a}}ron van~den Oord, Yazhe Li, and Oriol Vinyals.
\newblock Representation learning with contrastive predictive coding.
\newblock {\em CoRR}, abs/1807.03748, 2018.

\bibitem{bachman2019learning}
Philip Bachman, R.~Devon Hjelm, and William Buchwalter.
\newblock Learning representations by maximizing mutual information across views.
\newblock In Hanna~M. Wallach, Hugo Larochelle, Alina Beygelzimer, Florence d'Alch{\'{e}}{-}Buc, Emily~B. Fox, and Roman Garnett, editors, {\em Advances in Neural Information Processing Systems 32: Annual Conference on Neural Information Processing Systems 2019, NeurIPS 2019, December 8-14, 2019, Vancouver, BC, Canada}, pages 15509--15519, 2019.

\bibitem{khosla2021supervised}
Prannay Khosla, Piotr Teterwak, Chen Wang, Aaron Sarna, Yonglong Tian, Phillip Isola, Aaron Maschinot, Ce~Liu, and Dilip Krishnan.
\newblock Supervised contrastive learning.
\newblock In Hugo Larochelle, Marc'Aurelio Ranzato, Raia Hadsell, Maria{-}Florina Balcan, and Hsuan{-}Tien Lin, editors, {\em Advances in Neural Information Processing Systems 33: Annual Conference on Neural Information Processing Systems 2020, NeurIPS 2020, December 6-12, 2020, virtual}, 2020.

\bibitem{caruana1997multitask}
Rich Caruana.
\newblock Multitask learning.
\newblock {\em Mach. Learn.}, 28(1):41--75, 1997.

\bibitem{romera2013multilinear}
Bernardino Romera{-}Paredes, Hane Aung, Nadia Bianchi{-}Berthouze, and Massimiliano Pontil.
\newblock Multilinear multitask learning.
\newblock In {\em Proceedings of the 30th International Conference on Machine Learning, {ICML} 2013, Atlanta, GA, USA, 16-21 June 2013}, volume~28 of {\em {JMLR} Workshop and Conference Proceedings}, pages 1444--1452. JMLR.org, 2013.

\bibitem{luong2015multi}
Minh{-}Thang Luong, Quoc~V. Le, Ilya Sutskever, Oriol Vinyals, and Lukasz Kaiser.
\newblock Multi-task sequence to sequence learning.
\newblock In Yoshua Bengio and Yann LeCun, editors, {\em 4th International Conference on Learning Representations, {ICLR} 2016, San Juan, Puerto Rico, May 2-4, 2016, Conference Track Proceedings}, 2016.

\bibitem{lin2019pareto}
Xi~Lin, Hui{-}Ling Zhen, Zhenhua Li, Qingfu Zhang, and Sam Kwong.
\newblock Pareto multi-task learning.
\newblock In Hanna~M. Wallach, Hugo Larochelle, Alina Beygelzimer, Florence d'Alch{\'{e}}{-}Buc, Emily~B. Fox, and Roman Garnett, editors, {\em Advances in Neural Information Processing Systems 32: Annual Conference on Neural Information Processing Systems 2019, NeurIPS 2019, December 8-14, 2019, Vancouver, BC, Canada}, pages 12037--12047, 2019.

\bibitem{liu2020self}
Fangyu Liu, Ehsan Shareghi, Zaiqiao Meng, Marco Basaldella, and Nigel Collier.
\newblock Self-alignment pretraining for biomedical entity representations.
\newblock In Kristina Toutanova, Anna Rumshisky, Luke Zettlemoyer, Dilek Hakkani{-}T{\"{u}}r, Iz~Beltagy, Steven Bethard, Ryan Cotterell, Tanmoy Chakraborty, and Yichao Zhou, editors, {\em Proceedings of the 2021 Conference of the North American Chapter of the Association for Computational Linguistics: Human Language Technologies, {NAACL-HLT} 2021, Online, June 6-11, 2021}, pages 4228--4238. Association for Computational Linguistics, 2021.

\bibitem{wang2024dolphcoder}
Yejie Wang, Keqing He, Guanting Dong, Pei Wang, Weihao Zeng, Muxi Diao, Yutao Mou, Mengdi Zhang, Jingang Wang, Xunliang Cai, et~al.
\newblock Dolphcoder: Echo-locating code large language models with diverse and multi-objective instruction tuning.
\newblock {\em arXiv preprint arXiv:2402.09136}, 2024.

\bibitem{NQ}
Tom Kwiatkowski, Jennimaria Palomaki, Olivia Redfield, Michael Collins, Ankur~P. Parikh, Chris Alberti, Danielle Epstein, Illia Polosukhin, Jacob Devlin, Kenton Lee, Kristina Toutanova, Llion Jones, Matthew Kelcey, Ming{-}Wei Chang, Andrew~M. Dai, Jakob Uszkoreit, Quoc Le, and Slav Petrov.
\newblock Natural questions: a benchmark for question answering research.
\newblock {\em Trans. Assoc. Comput. Linguistics}, 7:452--466, 2019.

\bibitem{TriviaQA}
Mandar Joshi, Eunsol Choi, Daniel~S. Weld, and Luke Zettlemoyer.
\newblock Triviaqa: {A} large scale distantly supervised challenge dataset for reading comprehension.
\newblock In Regina Barzilay and Min{-}Yen Kan, editors, {\em Proceedings of the 55th Annual Meeting of the Association for Computational Linguistics, {ACL} 2017, Vancouver, Canada, July 30 - August 4, Volume 1: Long Papers}, pages 1601--1611. Association for Computational Linguistics, 2017.

\bibitem{HotpotQA}
Zhilin Yang, Peng Qi, Saizheng Zhang, Yoshua Bengio, William~W. Cohen, Ruslan Salakhutdinov, and Christopher~D. Manning.
\newblock Hotpotqa: {A} dataset for diverse, explainable multi-hop question answering.
\newblock In Ellen Riloff, David Chiang, Julia Hockenmaier, and Jun'ichi Tsujii, editors, {\em Proceedings of the 2018 Conference on Empirical Methods in Natural Language Processing, Brussels, Belgium, October 31 - November 4, 2018}, pages 2369--2380. Association for Computational Linguistics, 2018.

\bibitem{webqsp}
Wen{-}tau Yih, Matthew Richardson, Christopher Meek, Ming{-}Wei Chang, and Jina Suh.
\newblock The value of semantic parse labeling for knowledge base question answering.
\newblock In {\em Proceedings of the 54th Annual Meeting of the Association for Computational Linguistics, {ACL} 2016, August 7-12, 2016, Berlin, Germany, Volume 2: Short Papers}. The Association for Computer Linguistics, 2016.

\bibitem{DBLP:conf/nips/instructgpt}
Long Ouyang, Jeffrey Wu, Xu~Jiang, Diogo Almeida, Carroll~L. Wainwright, Pamela Mishkin, Chong Zhang, Sandhini Agarwal, Katarina Slama, Alex Ray, John Schulman, Jacob Hilton, Fraser Kelton, Luke Miller, Maddie Simens, Amanda Askell, Peter Welinder, Paul~F. Christiano, Jan Leike, and Ryan Lowe.
\newblock Training language models to follow instructions with human feedback.
\newblock In {\em NeurIPS}, 2022.

\bibitem{achiam2023gpt}
OpenAI.
\newblock {GPT-4} technical report.
\newblock {\em CoRR}, abs/2303.08774, 2023.

\bibitem{touvron2023LLaMA}
Hugo Touvron, Louis Martin, Kevin Stone, Peter Albert, Amjad Almahairi, Yasmine Babaei, Nikolay Bashlykov, Soumya Batra, Prajjwal Bhargava, Shruti Bhosale, Dan Bikel, Lukas Blecher, Cristian Canton{-}Ferrer, Moya Chen, Guillem Cucurull, David Esiobu, Jude Fernandes, Jeremy Fu, Wenyin Fu, Brian Fuller, Cynthia Gao, Vedanuj Goswami, Naman Goyal, Anthony Hartshorn, Saghar Hosseini, Rui Hou, Hakan Inan, Marcin Kardas, Viktor Kerkez, Madian Khabsa, Isabel Kloumann, Artem Korenev, Punit~Singh Koura, Marie{-}Anne Lachaux, Thibaut Lavril, Jenya Lee, Diana Liskovich, Yinghai Lu, Yuning Mao, Xavier Martinet, Todor Mihaylov, Pushkar Mishra, Igor Molybog, Yixin Nie, Andrew Poulton, Jeremy Reizenstein, Rashi Rungta, Kalyan Saladi, Alan Schelten, Ruan Silva, Eric~Michael Smith, Ranjan Subramanian, Xiaoqing~Ellen Tan, Binh Tang, Ross Taylor, Adina Williams, Jian~Xiang Kuan, Puxin Xu, Zheng Yan, Iliyan Zarov, Yuchen Zhang, Angela Fan, Melanie Kambadur, Sharan Narang, Aur{\'{e}}lien Rodriguez, Robert Stojnic, Sergey Edunov,
  and Thomas Scialom.
\newblock Llama 2: Open foundation and fine-tuned chat models.
\newblock {\em CoRR}, abs/2307.09288, 2023.

\bibitem{LLaMA3}
Meta.
\newblock Introducing meta llama 3: The most capable openly available llm to date, 2024.

\bibitem{qwen}
Jinze Bai, Shuai Bai, Yunfei Chu, Zeyu Cui, Kai Dang, Xiaodong Deng, Yang Fan, Wenbin Ge, Yu~Han, Fei Huang, Binyuan Hui, Luo Ji, Mei Li, Junyang Lin, Runji Lin, Dayiheng Liu, Gao Liu, Chengqiang Lu, Keming Lu, Jianxin Ma, Rui Men, Xingzhang Ren, Xuancheng Ren, Chuanqi Tan, Sinan Tan, Jianhong Tu, Peng Wang, Shijie Wang, Wei Wang, Shengguang Wu, Benfeng Xu, Jin Xu, An~Yang, Hao Yang, Jian Yang, Shusheng Yang, Yang Yao, Bowen Yu, Hongyi Yuan, Zheng Yuan, Jianwei Zhang, Xingxuan Zhang, Yichang Zhang, Zhenru Zhang, Chang Zhou, Jingren Zhou, Xiaohuan Zhou, and Tianhang Zhu.
\newblock Qwen technical report.
\newblock {\em arXiv preprint arXiv:2309.16609}, 2023.

\bibitem{youdao_bcembedding_2023}
Inc. NetEase~Youdao.
\newblock Bcembedding: Bilingual and crosslingual embedding for rag.
\newblock \url{https://github.com/netease-youdao/BCEmbedding}, 2023.

\bibitem{santhanam2022colbertv2}
Keshav Santhanam, Omar Khattab, Jon Saad{-}Falcon, Christopher Potts, and Matei Zaharia.
\newblock Colbertv2: Effective and efficient retrieval via lightweight late interaction.
\newblock In Marine Carpuat, Marie{-}Catherine de~Marneffe, and Iv{\'{a}}n Vladimir~Meza Ru{\'{\i}}z, editors, {\em Proceedings of the 2022 Conference of the North American Chapter of the Association for Computational Linguistics: Human Language Technologies, {NAACL} 2022, Seattle, WA, United States, July 10-15, 2022}, pages 3715--3734. Association for Computational Linguistics, 2022.

\bibitem{yuan2023rrhf}
Zheng Yuan, Hongyi Yuan, Chuanqi Tan, Wei Wang, Songfang Huang, and Fei Huang.
\newblock {RRHF:} rank responses to align language models with human feedback without tears.
\newblock {\em CoRR}, abs/2304.05302, 2023.

\bibitem{zhou2024lima}
Chunting Zhou, Pengfei Liu, Puxin Xu, Srinivasan Iyer, Jiao Sun, Yuning Mao, Xuezhe Ma, Avia Efrat, Ping Yu, Lili Yu, Susan Zhang, Gargi Ghosh, Mike Lewis, Luke Zettlemoyer, and Omer Levy.
\newblock {LIMA:} less is more for alignment.
\newblock In Alice Oh, Tristan Naumann, Amir Globerson, Kate Saenko, Moritz Hardt, and Sergey Levine, editors, {\em Advances in Neural Information Processing Systems 36: Annual Conference on Neural Information Processing Systems 2023, NeurIPS 2023, New Orleans, LA, USA, December 10 - 16, 2023}, 2023.

\bibitem{lu2023instag}
Keming Lu, Hongyi Yuan, Zheng Yuan, Runji Lin, Junyang Lin, Chuanqi Tan, Chang Zhou, and Jingren Zhou.
\newblock {\#}instag: Instruction tagging for analyzing supervised fine-tuning of large language models.
\newblock {\em CoRR}, abs/2308.07074, 2023.

\bibitem{dong2023abilities}
Guanting Dong, Hongyi Yuan, Keming Lu, Chengpeng Li, Mingfeng Xue, Dayiheng Liu, Wei Wang, Zheng Yuan, Chang Zhou, and Jingren Zhou.
\newblock How abilities in large language models are affected by supervised fine-tuning data composition.
\newblock {\em arXiv preprint arXiv:2310.05492}, 2023.

\bibitem{wu2024LLaMA}
Chengyue Wu, Yukang Gan, Yixiao Ge, Zeyu Lu, Jiahao Wang, Ye~Feng, Ping Luo, and Ying Shan.
\newblock Llama pro: Progressive llama with block expansion.
\newblock {\em arXiv preprint arXiv:2401.02415}, 2024.

\bibitem{dou2023art}
Shihan Dou, Enyu Zhou, Yan Liu, Songyang Gao, Jun Zhao, Wei Shen, Yuhao Zhou, Zhiheng Xi, Xiao Wang, Xiaoran Fan, et~al.
\newblock The art of balancing: Revolutionizing mixture of experts for maintaining world knowledge in language model alignment.
\newblock {\em arXiv preprint arXiv:2312.09979}, 2023.

\bibitem{tang2024mathscale}
Zhengyang Tang, Xingxing Zhang, Benyou Wang, and Furu Wei.
\newblock Mathscale: Scaling instruction tuning for mathematical reasoning, 2024.

\bibitem{dong2024self}
Guanting Dong, Keming Lu, Chengpeng Li, Tingyu Xia, Bowen Yu, Chang Zhou, and Jingren Zhou.
\newblock Self-play with execution feedback: Improving instruction-following capabilities of large language models.
\newblock {\em arXiv preprint arXiv:2406.13542}, 2024.

\bibitem{UniK-QA}
Barlas Oguz, Xilun Chen, Vladimir Karpukhin, Stan Peshterliev, Dmytro Okhonko, Michael Schlichtkrull, Sonal Gupta, Yashar Mehdad, and Scott Yih.
\newblock {U}ni{K}-{QA}: Unified representations of structured and unstructured knowledge for open-domain question answering.
\newblock In {\em Findings of the Association for Computational Linguistics: NAACL 2022}, pages 1535--1546, Seattle, United States, July 2022. Association for Computational Linguistics.

\bibitem{dong2023bridging}
Guanting Dong, Rumei Li, Sirui Wang, Yupeng Zhang, Yunsen Xian, and Weiran Xu.
\newblock Bridging the kb-text gap: Leveraging structured knowledge-aware pre-training for kbqa.
\newblock In {\em Proceedings of the 32nd ACM International Conference on Information and Knowledge Management}, pages 3854--3859, 2023.

\bibitem{luo2023chatkbqa}
Haoran Luo, Zichen Tang, Shiyao Peng, Yikai Guo, Wentai Zhang, Chenghao Ma, Guanting Dong, Meina Song, Wei Lin, et~al.
\newblock Chatkbqa: A generate-then-retrieve framework for knowledge base question answering with fine-tuned large language models.
\newblock {\em arXiv preprint arXiv:2310.08975}, 2023.

\bibitem{karpukhin2020dense}
Vladimir Karpukhin, Barlas Oğuz, Sewon Min, Patrick Lewis, Ledell Wu, Sergey Edunov, Danqi Chen, and Wen tau Yih.
\newblock Dense passage retrieval for open-domain question answering, 2020.

\bibitem{Wikidata}
Denny Vrande\v{c}i\'{c} and Markus Kr\"{o}tzsch.
\newblock Wikidata: A free collaborative knowledgebase.
\newblock {\em Commun. ACM}, 57(10):78–85, sep 2014.

\bibitem{loshchilov2017decoupled}
Ilya Loshchilov and Frank Hutter.
\newblock Decoupled weight decay regularization.
\newblock {\em arXiv preprint arXiv:1711.05101}, 2017.

\bibitem{zheng2024LLaMAfactory}
Yaowei Zheng, Richong Zhang, Junhao Zhang, Yanhan Ye, Zheyan Luo, and Yongqiang Ma.
\newblock Llamafactory: Unified efficient fine-tuning of 100+ language models.
\newblock {\em arXiv preprint arXiv:2403.13372}, 2024.

\bibitem{bai2023qwen}
Jinze Bai, Shuai Bai, Yunfei Chu, Zeyu Cui, Kai Dang, Xiaodong Deng, Yang Fan, Wenbin Ge, Yu~Han, Fei Huang, Binyuan Hui, Luo Ji, Mei Li, Junyang Lin, Runji Lin, Dayiheng Liu, Gao Liu, Chengqiang Lu, Keming Lu, Jianxin Ma, Rui Men, Xingzhang Ren, Xuancheng Ren, Chuanqi Tan, Sinan Tan, Jianhong Tu, Peng Wang, Shijie Wang, Wei Wang, Shengguang Wu, Benfeng Xu, Jin Xu, An~Yang, Hao Yang, Jian Yang, Shusheng Yang, Yang Yao, Bowen Yu, Hongyi Yuan, Zheng Yuan, Jianwei Zhang, Xingxuan Zhang, Yichang Zhang, Zhenru Zhang, Chang Zhou, Jingren Zhou, Xiaohuan Zhou, and Tianhang Zhu.
\newblock Qwen technical report.
\newblock {\em CoRR}, abs/2309.16609, 2023.

\bibitem{jiang2023mistral}
Albert~Q. Jiang, Alexandre Sablayrolles, Arthur Mensch, Chris Bamford, Devendra~Singh Chaplot, Diego de~Las~Casas, Florian Bressand, Gianna Lengyel, Guillaume Lample, Lucile Saulnier, L{\'{e}}lio~Renard Lavaud, Marie{-}Anne Lachaux, Pierre Stock, Teven~Le Scao, Thibaut Lavril, Thomas Wang, Timoth{\'{e}}e Lacroix, and William~El Sayed.
\newblock Mistral 7b.
\newblock {\em CoRR}, abs/2310.06825, 2023.

\bibitem{javaheripi2023phi}
Mojan Javaheripi, S{\'e}bastien Bubeck, Marah Abdin, Jyoti Aneja, Sebastien Bubeck, Caio C{\'e}sar~Teodoro Mendes, Weizhu Chen, Allie Del~Giorno, Ronen Eldan, Sivakanth Gopi, et~al.
\newblock Phi-2: The surprising power of small language models.
\newblock {\em Microsoft Research Blog}, 2023.

\end{thebibliography}

\clearpage
\appendix
\begin{center}
{\Large \textbf{Appendix}}
\end{center}


\setcounter{section}{0}
\renewcommand{\thesection}{\Alph{section}}




\tableofcontents
\clearpage

\section{More Details about DPA-RAG}
\label{app:More Details about DPA-RAG}

\subsection{The Overall Algorithm Workflow of DPA-RAG}

In this section, we delve into the overall workflow of the DPA-RAG algorithm, which can be divided into \textbf{Reranker Training Algorithm }and \textbf{LLM-based Generator Training}.


\textbf{Reranker Training Algorithm:} Given the train set $ \widetilde{D}_{\text{train}} = \{q_{i},D_{q_{i}}, y_{q_{i}} \}_{i=1}^{N_{\text{train}}}$, we initially perform preference knowledge mining techniques to select, augment and filter the data to construct a preference-aligned dataset $\widetilde{D}_{\text{pref}}$. Subsequently, relying on the $\widetilde{D}_{\text{pref}}$, we perform multi-grained distillation alignments with MGDA-UB stategy to better fine-tune a preference-aligned reranker. The detailed process is listed in algorithm diagram \ref{reranker_training}.

\textbf{LLM-based Reader Training Algorithm:} As shown in algorithm diagram \ref{llm_training}, for open-source LLM-based reader, we directly utilize the preference-aligned reranker to perform preference-based reranking on retrieved documents in $\widetilde{D}_{\text{train}}$\footnote{The training set $\widetilde{D}_{\text{train}}$ consists of the original training set $\widetilde{D}_{\text{train}}^{ori}$ and ${\widetilde{D}_{\text{aug}} \in \widetilde{D}_{\text{pref}}}$ with five query augmentations.} and $\widetilde{D}_{\text{test}}$, resulting in sorted datasets $\widetilde{D}_{\text{train}}^{\text{rank}}$ and $\widetilde{D}_{\text{test}}^{\text{rank}}$. In addition, we also construct a dataset $\widetilde{D}_{\text{train}}^{\text{PA}}$ for the knowledge self-alignment task based on $\widetilde{D}_{\text{pref}}$. Initially, we use $\widetilde{D}_{\text{train}}^{\text{PA}}$ for the pre-aligned task, then we load the pre-trained model parameters and then conduct vanilla QA supervised fine-tuning based on $\widetilde{D}_{\text{train}}^{\text{rank}}$. During the inference phase, we input the preference-sorted test set $\widetilde{D}_{\text{test}}^{rank}$ into the LLM to complete the prediction.

For close-source LLM-based reader, the process is more simple: the preference-aligned reranker is used to sort documents in the test set $\widetilde{D}_{\text{test}} \to \widetilde{D}_{\text{test}}^{\text{rank}}$, then we use LLMs for the prediction process.

\begin{algorithm*}
\caption{Reranker Training}
\label{reranker_training}
\begin{algorithmic}[1]

\Procedure{ConstructPreferenceDataset}{$\widetilde{D}_{\text{train}}$}.
    \State $\widetilde{D}_{\text{pref}} \gets \emptyset$
    \State From $(q_i, D_{q_i}, y_{q_i}) \in \widetilde{D}_{\text{train}}$, we select the $\widetilde{D}_{\text{sub}} = \{q_{i}, D_{q_{i}}^{\text{sub}}, Y_{i}^{\text{sub}}\}_{i = 1}^{N}$.
    \ForAll{$ \{q_{i}, D_{q_{i}}^{\text{sub}}, Y_{i}^{\text{sub}}\} \in 
    \widetilde{D}_{\text{sub}}$} \Comment{Mine Preference Knowledge}

        \ForAll {$\{d_{i} | i = 1, 25, 50, 100\} \in D_{q_{i}}^{\text{sub}}$}
            \State $a_{\text{LLM}} \gets \text{LLM answer to query } q_i$
            \State $a_{\text{docs}} \gets \text{Correct answer from } d_{i}$
            \If{\(a_{\text{LLM}} \neq y_n\) \text{ and } \(a_{\text{docs}} = y_n\)}
                \State $\widetilde{D}_{\text{pref}} \gets \widetilde{D}_{\text{pref}} \cup \{(q_{i}, D_{q_{i}}^{\text{sub}}, Y_{i}^{\text{sub}})\}$ \Comment{Aligned Knowledge}
                \State Continue
            \ElsIf{\(a_{\text{LLM}} = y_n\) \text{ and } \(a_{\text{docs}} \neq y_n\)}
                \State $\widetilde{D}_{\text{pref}} \gets \widetilde{D}_{\text{pref}} \cup \{(q_{i}, D_{q_{i}}^{\text{sub}}, Y_{i}^{\text{sub}})\}$ \Comment{Unaligned Knowledge}
                \State Continue
            \EndIf
            \EndFor
        \EndFor
    \State $G_{\theta} \gets \text{Augmented query generator}$
    \State $R \gets \{\text{Complexity, Constraint, SPARQL, Decomposition, Rephrasing}\}$
    \ForAll{$R_i$ in $R$}
        \ForAll{$(q_i, D_{q_i}) \in \widetilde{D}_{\text{pref}}$}
            \State $q_{\text{aug}, i} \gets G_{\theta}(R_i, q_i, D_{q_i})$
            \State $D_{r_i} \gets D_{r_i} \cup \{(q_{\text{aug}, i}, D_{q_i}, y_{q_i})\}$
        \EndFor
        \State $\widetilde{D}_{\text{pref}} \gets \widetilde{D}_{\text{pref}} \cup \left( \cup_{i=1}^{n} D_{r_i} \right)$
    \EndFor
    \State $p_{\Theta} \gets \text{NLI model for quality filtering}$
    \ForAll{augmented query $q_{\text{aug}}$ in $\widetilde{D}_{\text{pref}}$}
        \State $score_{\theta} \gets p_{\Theta}(q, q_{\text{aug}})$
        \If{\(score_{\theta} \text{ is not ``entailment''}\)}
            \State $\widetilde{D}_{\text{pref}} \gets \widetilde{D}_{\text{pref}} \setminus \{(q_{\text{aug}}, D_{q_i}, y_{q_i})\}$
        \EndIf
    \EndFor
    \State \textbf{return} $\widetilde{D}_{\text{pref}}$
\EndProcedure

\Procedure{MultiGrainedDistillationAlignment}{$\widetilde{D}_{\text{pref}}$}
    \State Initialize model parameters $\theta^{sh}, \theta^1, \ldots, \theta^T$
    \Repeat
        \State Compute losses $\mathcal{L}_{\text{CPD}}$, $\mathcal{L}_{\text{FPR}}$, $\mathcal{L}_{\text{SCA}}$
        \Procedure{MGDA-UB}{$\theta^{sh}, \theta^1, \ldots, \theta^T, c^t$}
            \State $\mathbf{Z} \gets \sum_{t=1}^{T} c^t \nabla_{\theta^{sh}} \hat{\mathcal{L}}^{t}(\theta^{sh}, \theta^t)$
            \State Optimize MTL weights $\alpha^t$ for Pareto optimal solution
            \State $\mathbf{L} \gets \sum_{t=1}^{T} c^t \hat{\mathcal{L}}^{t}(\theta^{sh}, \theta^t)$
            \State \textbf{return} $\mathbf{L}$
        \EndProcedure
        \State Update model parameters $\theta^{sh}, \theta^1, \ldots, \theta^T$ to minimize $\mathbf{L}$
    \Until{convergence}
    \State \textbf{return} Optimized parameters $\theta^{sh}, \theta^1, \ldots, \theta^T$

\EndProcedure

\end{algorithmic}
\end{algorithm*}

\begin{algorithm*}
\caption{LLM-based Reader Training}
\label{llm_training}
\begin{algorithmic}[1]

\Procedure{Pre-Align}{$\widetilde{D}_{\text{pref}}$, $k$}
    \ForAll{$\{q_{i}, D_{\text{pref}}, y_{q_{i}}\} \in \widetilde{D}_{\text{pref}}$}
        \State Select one document from $D_{\text{pref}}$
        \State Randomly select $k-1$ documents from $D = \{d_i\}^{N}_{i=1}$
        \State Construct Top-k document set $D_{\text{align}} = \{d_{\text{pref}}, d_{\text{rand}_1}, \ldots, d_{\text{rand}_{k-1}}\}$
        \State Initialize prompt with the selected documents and query
    \EndFor
    \State Fine-tune the LLMs with the objective $\mathcal{L}\left(\theta\right) = {\sum\limits_{(q_{i},D_{\text{align}}, y_{q_i}) \in \mathcal{D} }{\log{\mathbf{P}_{\theta}\left(y_{q_i} | prompt(q_{i},D_{\text{align}}) \right)}}}$

\EndProcedure

\Procedure{Supervised Fine-Tuning}{$\mathcal{D}$, Pre-Aligned Parameters}
    \State Load pre-warmed parameters from PreAligned stage
    \State Merge augmented dataset as $\widetilde{D}_{\text{train}} = \widetilde{D}_{\text{train}} \cup ( \cup_{i=1}^{n} \widetilde{D}_{r_i} )$
    \ForAll{$\{q_{i}, D_{q_{i}}, y_{q_{i}}\} \in \widetilde{D}_{\text{train}}$}
        \State $D_{q_i}^{\text{rank}} \gets \text{Top-K} \ [ \ \text{Reranker}(q_{i}, D_{q_{i}}) \ ]$
        \State $\widetilde{D}_{\text{train}}^{\text{rank}} \gets \{(q_{i}, D_{q_i}^{\text{rank}}, y_{q_{i}})\}$
    \EndFor
    \State Perform supervised fine-tuning
\EndProcedure
\end{algorithmic}
\end{algorithm*}

\subsection{Preference Scoring Mechanism for Different LLMs}
\label{app:score}

In practice, we find that models with fewer than 7B parameters struggle with instruction-following capabilities, making it difficult for them to perform the scoring task. To address this, we follow the RankLLaMA~\citep{ma2023finetuning} and RePLUG~\citep{shi2023replug}, utilizing the output's logit as the basis for scoring as follow:
\begin{equation}
    r_{\theta}(q, d_i) =  {\log{\mathbf{P}_{\theta}\left(\ \text{prompt}\ (q, d_i)  \right)}}
\end{equation}
\begin{equation}
\label{eq17}
s_i = a \cdot r_{\theta}(q, p_i) + (1-a) \cdot s_{R}(q, p_i)
\end{equation}
where $q$, $d_i$ denotes the query and top i-th document. $ \log \mathbf{P}(\cdot )$ represents the model's probability distribution. $\text{Prompt}$ denotes the prompt template. $s_i$ is the final preference score of i-th retrieved document. For the hyper-parameter $a$, we follow QLM Reranker~\citep{zhuang2023open} and set it to 0.8 without performing any grid search. Next, we rank them to obtain the preference order $\{o_1, o_2,..,o_n \mid r_{\theta}, s_R\}$ according to $  \{s_{i}\}_{i=1}^{K}$.

For the 7B and 13B models, we observe that these models fundamentally possess the capability to follow instructions in our preliminary experiments. Therefore, we ask them to perform preference scores from 1 to 5. Then we normalize the preference score $r_{\theta}(q, d_i)$ and sum it with the retriever's similarity score $s_{R}(q, d_i)$ as equation \ref{eq17}. Finally, we rank them to obtain the preference order.

As the result in Table \ref{main-results}, for powerful LLMs (such as GPT-3.5 and GPT-4), we find that a pairwise comparative ranking can achieve a more precise preference ordering compared to the ranking by scoring each paragraph individually. Therefore, we perform $C_{k}^2$ pair-wise comparisons of knowledge documents as PRP~\citep{qin2024large} through LLMs to obtain the preference ordering results.

\section{More Details on Experiment Setup}
\label{app:setup}

\subsection{Datasets}

In this section, we report the detailed information of our 4 datasets, including NaturalQuestions (NQ), TriviaQA (TQA), HotpotQA (HQA), WebQuestionsSP (WebQSP). 

\textbf{Natural Questions (NQ)}~\citep{NQ} dataset, with its approximately 300,000 real Google searches and corresponding answers from Wikipedia, annotated for detailed context and brief replies, is crucial for developing question-answering systems, enhancing AI's comprehension of natural language.

\textbf{TriviaQA (TQA)}~\citep{TriviaQA} serves as a benchmark for QA models, with its extensive set of over 650,000 question-answer pairs sourced from quizzes and trivia competitions. Each question is linked to supporting documents, presenting a challenge for systems to extract correct information from various subjects, which in turn evaluates their information gathering and language comprehension capabilities.

\textbf{HotpotQA (HQA)}~\citep{HotpotQA} dataset comprises 113,000 questions necessitating answers through multi-step logic. It pushes the envelope in AI development by demanding linkage of several documents for inferencing comprehensive answers, aiming to improve AI abilities in complex understanding far exceeding simple fact extraction.

\textbf{WebQuestionsSP (WebQSP)}~\citep{webqsp} dataset consists of more than 4,700 Google Suggest-derived questions, each associated with a query in SPARQL format that retrieves answers from the Freebase. It is specifically crafted for refining QA systems' semantic parsing skills and their ability to transform natural language into formal database queries, thereby pushing the boundaries of AI in processing and understanding intricate queries from real-life scenarios.

\subsection{Prompt Templates}

In the vanilla SFT stage, we follow the template of the RA-Judgement~\citep{ren2023investigating} as follow:

\begin{tcolorbox}[
colback=white!10!white,
colframe=black!75!black,
title=Prompt Template of SFT Stage,
breakable]
Given the documents \{Top-K Documents\}. Answer the following question based on the given information or your internal knowledge with one or few words without the source. Query: \{Query\}. 

\end{tcolorbox}

For the pre-aligned stage, our prompt template is almost aligned with the SFT stage template. The only difference is that we add an additional judgment statement that allows LLMs to distinguish whether the influence of the preference document $d_{q}$ on answering questions is positive or negative, thus implicitly learning the ability to distinguish between aligned knowledge and unaligned knowledge. The prompt template is displayed as follow:

\begin{tcolorbox}[
colback=white!10!white,
colframe=black!75!black,
title=Prompt Template of Pre-aligned Stage,
breakable]
Given the documents $\{D_{\text{align}}= (d_{q}, d_{{\text{rand}}_1}, \ldots, d_{{\text{rand}}_{k-1}})\}$. Answer the following question based on the given information or your internal knowledge with few words without the source. Query: $\{q\}$. \\
$[\text{Judgement}]$ document-$\{i_{d_{q}}\}$ is Positive or Negative knowledge for answering question.
\end{tcolorbox}

where $d_{q}$ denotes the preference document that influences the LLM's reasoning results for query $q$. $\{d_{{\text{rand}}_1}, \ldots, d_{{\text{rand}}_{k-1}}\}$ denotes $k-1$ random documents from the retrieved corpus $D_{\text{align}}$. Moreover, $i_{d_q}$ denotes the order of $d_q$ in $D_{\text{align}}$.

For data augmentation process, motivated by the data augmentation process of several works~\citep{luo2023wizardmath,luo2023wizardcoder,yu2023metamath,yuan2023scaling,dong2024self,li2023query}, we employ \texttt{gpt-3.5-turbo-0613} APIs with a temperature of 1.0. Then we specially design a augmentation prompt for RAG as follow:

\begin{tcolorbox}[
colback=white!10!white,
colframe=black!75!black,
title=Query Augmentation Prompt,
breakable]
You are an AI assistant helping me rewrite the query. I will give you the original query, reference document, title and rewriting requirements. Please rewrite the query based on the following information:\\

\textbf{Original Query}: \{Query\}\\
\textbf{Reference Documents}: \{Top-K Documents\}\\
\textbf{Title}: \{Title\}\\
\textbf{Augmentation Requirements}: \{Augmneted Requirements\}\\
\textbf{New Queries:}
\label{tab:query-aug prompt}
\end{tcolorbox}

\subsection{Implementation Details}
Here, we report our detailed information of DPA-RAG, as a retriever-reranker-reader architecture:

For the retriever, following previous works~\citep{UniK-QA,dong2023bridging,luo2023chatkbqa}, we utilize Dense Document Retriever (DPR)~\citep{karpukhin2020dense} for encoding documents and questions respectively. After that, we use it retrieves the top 100 relevant Wikipedia documents~\citep{Wikidata} according to the dot-product similarity.

For the reranker, we use the BGE~\citep{bge_embedding} as our backbone model. Specifically, we adjust our batch size to 16. We fine-tune our reranker for 10 epochs and set the learning rate to 1e-5. We utilize the BGE reranker to order the top 100 retrieved documents to obtain the top-3 results.\footnote{we use mDeberta as our filtering model, which can be downloaded at \url{https://huggingface.co/MoritzLaurer/mDeBERTa-v3-base-xnli-multilingual-nli-2mil7}}. 

For the QA fine-tuning setting,  we employ the AdamW optimizer~\citep{loshchilov2017decoupled} to train our LLMs for 3 epochs. Moreover, we set our training batch size to 128. We use eight A100 80g GPUs to fine-tune all models with top-3 documents. Our learning rate is set as 7e-5 with a 3\% warmup process. For all experiments, we conduct them using the LLaMA Factory framework~\citep{zheng2024LLaMAfactory} with the default system prompts of the model. We use the version 0.6.3\footnote{\url{https://github.com/hiyouga/LLaMA-Factory/releases/tag/v0.6.3}} for training LLaMA2, Mistral, Qwen1.5 and Phi2. In addition, we use the version 0.8.1 \footnote{\url{https://github.com/hiyouga/LLaMA-Factory/releases/tag/v0.8.1}} for Qwen2 and LLaMA3. We report the average performance from five experiments, each with a different random seed.

To facilitate the reproduction of our results, all datasets and evaluation benchmarks used in our experiments have been open-source, and their detailed sources are indicated. We promise to open-source our code after the blind review process.

\subsection{Baselines}

We mainly compare DPA-RAG with multiple strong baselines by using reranker-based methods and preference aligned methods for RAG as follow:

\textbf{Reranker-based Baselines:}
\begin{itemize}[leftmargin=1em]
\item \textbf{RankGPT}~\citep{sun2023chatgpt} leverages listwise prompting and utilizes specific distillation method to replicate the document re-ranking abilities of GPT-3.5 within a smaller ranking model.

\item \textbf{LRL}~\citep{ma2023zero} is a model that utilizes GPT-3.5 as a zero-shot reranker for listwise ranking, which directly generates a ranking list of candidate documents.

\item \textbf{PRP}~\citep{qin2024large}, Pairwise Ranking Prompting, which involves submitting a query alongside a pair of documents into the prompt, enabling large language models to perform ranking tasks.

\item \textbf{RankLLaMA}~\citep{ma2023finetuning}, based on LLaMA, is trained as a pointwise reranker. This approach involves passing both query and document together to the model. RankLLaMA generates a similarity score reflecting the document's relevance to the query.

\item \textbf{BGE}~\citep{bge_embedding} is a general Embedding Model developed by BAAI. The reranker use the cross-encoder structure to do full-attention on the input pair.

\item \textbf{BCEmbedding}~\citep{youdao_bcembedding_2023}, Bilingual and Crosslingual Embedding in English and Chinese, developed by NetEase Youdao. Their Reranker is particularly proficient at refining search results and improving ranking tasks.

\item \textbf{ColBERTv2}~\citep{santhanam2022colbertv2}, a model employs a combination of denoised supervision and residual compression techniques, utilizing token-level decomposition during late interaction.

\end{itemize}

\textbf{Preference-aligned Baselines:}
\begin{itemize}[leftmargin=1em]
\item \textbf{KnowPAT}~\citep{zhang2024knowledgeable} is a framework that constructs a knowledgeable preference set to align model preferences with knowledge. This framework effectively guides language models to select relevant knowledge for specific inquiries, enhancing their ability to provide pertinent information.

\item \textbf{REPLUG}~\citep{shi2023replug} It is a retrieval-enhanced language modeling framework that dynamically optimizes the retriever through the output probability of a black box large language model.

\item \textbf{RA-Judgement}~\citep{ren2023investigating}, which is known as Retrieval-augmented
judgement. In this work, authors explores the knowledge boundary problem of RAG and proposes two experimental settings, Priori Judgment and Posteriori Judgment. RA-judgment is a dynamic improvement method based on Priori Judgment, which can better capture factual information.

\item \textbf{RRHF}~\citep{yuan2023rrhf} is a training paradigm, which aims to align probabilities of model responses with human preferences by a ranking loss, which can retain the performance of Proximal Policy Optimization (PPO) and is much simpler. 

\item \textbf{RAFT}~\citep{zhang2024raft} boosts a language model's proficiency in answering questions within a specific domain by teaching it to disregard irrelevant documents and reference pertinent segments from retrieved texts. It enhances the model's reasoning capabilities and effectiveness in domain-related tasks while maintaining resilience against incorrect retrievals.

\item \textbf{FILCO}~\citep{wang2023learning} It is a data selection method based on vocabulary and information theory to improve the quality of generated answers provided to the generative model by filtering useful context in the training data.
\end{itemize}

Furthermore, We also provide a detailed introduction to the \textbf{LLM reader model} used by DPA-RAG:

\begin{itemize}[leftmargin=1em]
\item \textbf{LLaMA2}~\citep{touvron2023LLaMA} is an upgraded version of LLaMA developed by MetaAI. It utilizes more robust data cleaning and mixing techniques, and up-samples sources closest to factual information, which can enhance knowledge and reduce hallucinations. Additionally, it employs Grouped-Query Attention technology to lessen reliance on memory.

\item \textbf{LLaMA3}~\citep{LLaMA3}, created by MetaAI, the newest version of the LLaMA series, LLaMA3, includes major enhancements. In contrast to LLaMA2, LLaMA3 incorporates a larger training dataset, extended context length, and an enriched vocabulary, leading to better performance on a range of tasks. Additionally, LLaMA3 offers notable improvements in contextual comprehension and language generation, setting it apart from its predecessor.

\item \textbf{Qwen1.5}~\citep{bai2023qwen} series, created by Alibaba, comprises language models with advanced features like SwiGLU activation, attention QKV bias, group query attention, and a combination of sliding window and full attention mechanisms. These models boast robust fundamental abilities, particularly in language comprehension.

\item \textbf{Qwen2}~\citep{bai2023qwen}, developed by Alibaba, is available in several sizes: Qwen2-0.5B /1.5B /7B and 72B. This model is trained on data sources spanning 29 kinds of languages, enabling it to perform exceptionally well in multilingual tasks. Additionally, Qwen2 exhibits strong capabilities in coding and mathematics. Qwen2-72B-Instruct is notable for its ability to handle input windows of up to 128K tokens in length, making it exceptionally well-suited for processing long texts and tackling complex tasks.

\item \textbf{Mistral}~\citep{jiang2023mistral}, a language model boasting 7 billion parameters, is engineered by Mistral AI for exceptional performance and efficiency. Mistral 7B utilizes Packet Query Attention to accelerate inference and integrates Sliding Window Attention to efficiently manage sequences of varying lengths, all while minimizing inference costs.

\textbf{Phi2}~\citep{javaheripi2023phi}, proposed by Microsoft, is a powerful small language model with 2.7 billion parameters. Despite its relatively modest size, Phi-2 demonstrates exceptional reasoning and language comprehension capabilities. At its release, it showcased great performance among small foundational LLMs. In different benchmark tests, model's performance was comparable to, or even surpassed, models that are 25 times larger.

\item \textbf{GPT-3.5 and GPT-4}~\citep{achiam2023gpt}, proposed by OpenAI, which are part of the GPT families that incorporate a multi-step reinforcement learning from human feedback (RLHF) techniques. the algorithm not only enhances the models' instruction-following ability but also significantly reduces the likelihood of producing harmful or toxic content. Moreover, GPT-4 introduces support for image inputs and attains human-like performance on a range of benchmarks.
\end{itemize}

\section{More Details about Experimental Results}

\begin{table}[!t]
    \centering
    \small
    \renewcommand{\arraystretch}{1.2} 
    \setlength{\tabcolsep}{4mm} 
    \caption{Detailed Ablations of LLaMA2-7B on NQ and TQA. Point-wise., Pair-wise., CPA denotes Point-wise, Pair-wise and Contrastive Preference Alignment respectively.}
    
    \begin{tabular}{ccccc}
    \toprule
     \textbf{Method}   & \multicolumn{2}{c}{\textbf{NQ}} & \multicolumn{2}{c}{\textbf{TQA}} \\
    \cmidrule(lr){2-3}\cmidrule(lr){4-5}
        & \textbf{Hits@1} & \textbf{F1} & \textbf{Hits@1} & \textbf{F1} \\
    \midrule
    LLaMA2-7B DPA-RAG & 56.03 & 60.19 & 70.16 & 70.29 \\

\hline
\multicolumn{3}{c}{\textit{\textbf{Preference Knowledge Construction}}} \\
\quad w/o Query Aug. & -2.13 & -2.31 & -2.62 & -2.87 \\
w/o Filtering. & -0.92 & -0.71 & -1.39 & -1.45 \\

\hline
\multicolumn{3}{c}{\textit{\textbf{Multi-Grained Distillation Alignment}}} \\
    
\quad\quad \ w/o point-wise. & -1.95 & -2.12 & -2.43 & -2.43 \\
\quad\quad w/o pair-wise. & -0.98 & -0.92 & -1.51 & -1.74 \\
w/o CPA & -1.54 & -1.12 & -1.84 & -2.13 \\
\quad\quad\quad w/o MGDA-UB. & -0.52 & -0.77 & -0.84 & -1.10 \\

\hline
\multicolumn{2}{c}{\textit{\textbf{Knowledge Self-Alignment}}} \\

\quad w/o Pre-Align. & -1.72 & -1.76 & -2.21 & -2.45 \\
\bottomrule
    
    LLaMA2-7B RAG & 50.94 & 54.76 & 63.90 & 63.80 \\
    \bottomrule
    \end{tabular}
    \label{tab:ablation}

\end{table}

\subsection{Detailed Results for Ablation Studies}
\label{app:ablation}

Table \ref{tab:ablation} presents the detailed ablation results of our DPA-RAG across three key phases, with ``w/o'' indicating the model's version without a particular module. Our findings are as follows:

\begin{itemize}[leftmargin=1em]
\item DPA-RAG's result declines when any of its components are removed, further validating the necessity of each part we designed.

\item Focusing on the Preference Knowledge Construction stage, we notice that the Query Augmentation methods lead to a substantial improvement in performance, which is in line with our expectations. These strategies introduce additional supervision signals during the training stages of both the Reranker and the Reader, yielding a joint boost to the DPA-RAG framework. Moreover, the quality filtering process also brings slight performance gains, underscoring the importance of maintaining intent consistency between original and augmented data.

\item In the multi-grained distillation alignment stage, each task independently provides stable gains in both NQ and TQA. Point-wise preference alignment, as a fundmental capability for distinguishing knowledge preferences, brings the largest gains in aligning LLMs' preferences. Notably, the MGDA-UB strategy further yields stable gains on top of the joint optimization of three tasks, proving the necessity of introducing multi-task balance optimization.

\item The pre-aligned phase also shows steady performance gains, especially evident in TQA. In practice, we find that the potential for internal alignment in TQA is even greater than external, differing from NQ and HQA. Therefore, this insight also highlights the necessity of dual alignment to align datasets from different domains.
\end{itemize}

\subsection{Details about Diverse Query Augmentations}

\label{app:aug}

\paragraph{Case Study of Augmented Queries.} Table \ref{app:case-study} shows some samples which are generated by \texttt{gpt-3.5-turbo-0613} APIs 
in the way of different augmneted requierment, respectively. We can observe that the complexity level of the augmented data showcased in the case is generally consistent with the trend of complexity and diversity scores presented in Table \ref{tab:aug}.

\begin{table}[!t]
\centering
\small
\renewcommand{\arraystretch}{1.3}
\caption{Examples of different methods for generating new queries.}
\begin{tabular}{|c|m{3.2cm}|m{7.2cm}|}
\hline
\textbf{Method} &\centering\textbf{Requirement}  &\centering\arraybackslash \textbf{Query} \\
\hline
Origin &\centering - &\centering\arraybackslash What screenwriter with credits for ``Evolution'' co-wrote a film starring Nicolas Cage and Téa Leoni? \\
\hline
Rephrasing &\centering Rephrase the original query with the same intention. &\centering\arraybackslash Who is the screenwriter credited for ``Evolution'' who also co-authored a movie featuring Nicolas Cage and Téa Leoni? \\
\hline
Decomposition &\centering Decompose the original query into several sub-problems.  &\centering\arraybackslash Sub-problem 1: Identify the screenwriter who has credits for the film ``Evolution''. 
Sub-problem 2: Determine if the screenwriter from sub-problem 1 has also co-written a film where Nicolas Cage and Téa Leoni were cast. \\
\hline
SPARQL &\centering Rewrite the original query based on the SPARQL syntax and generate it directly.   & \begin{lstlisting}[language=SPARQL]
SELECT ?screenwriter WHERE {
  ?film rdf:type dbo:Film .
  ?film dbo:writer ?screenwriter .
  ?film dbo:starring dbr:Nicolas_Cage .
  ?film dbo:starring dbr:Tea_Leoni .
  ?screenwriter dbo:film dbr:Evolution .
  ?screenwriter rdfs:label ``David Weissman'' .
}
\end{lstlisting} \\
\hline
Constraint &\centering Add more conditional and constrained statements to the original query.    &\centering\arraybackslash Which screenwriter, known for working on the movie ``Evolution'', also co-authored a screenplay for a feature film that includes Nicolas Cage and Téa Leoni in the cast, and has a history of collaboration with David Diamond? \\
\hline
Complexity &\centering Increase the semantic complexity of the original query. &\centering\arraybackslash Which scriptwriter, known for his partnership with David Diamond and shared film credits on ``Evolution'', also co-authored a screenplay that featured Nicolas Cage and Téa Leoni in leading roles, after initially meeting his writing colleague at Akiba Hebrew Academy and making their screenwriting sale debut with ``The Whiz Kid'' to 20th Century Fox? \\
\hline
\end{tabular}

\label{app:case-study}
\end{table}

\begin{figure}[t]
\centering
\includegraphics[width=0.7\linewidth]{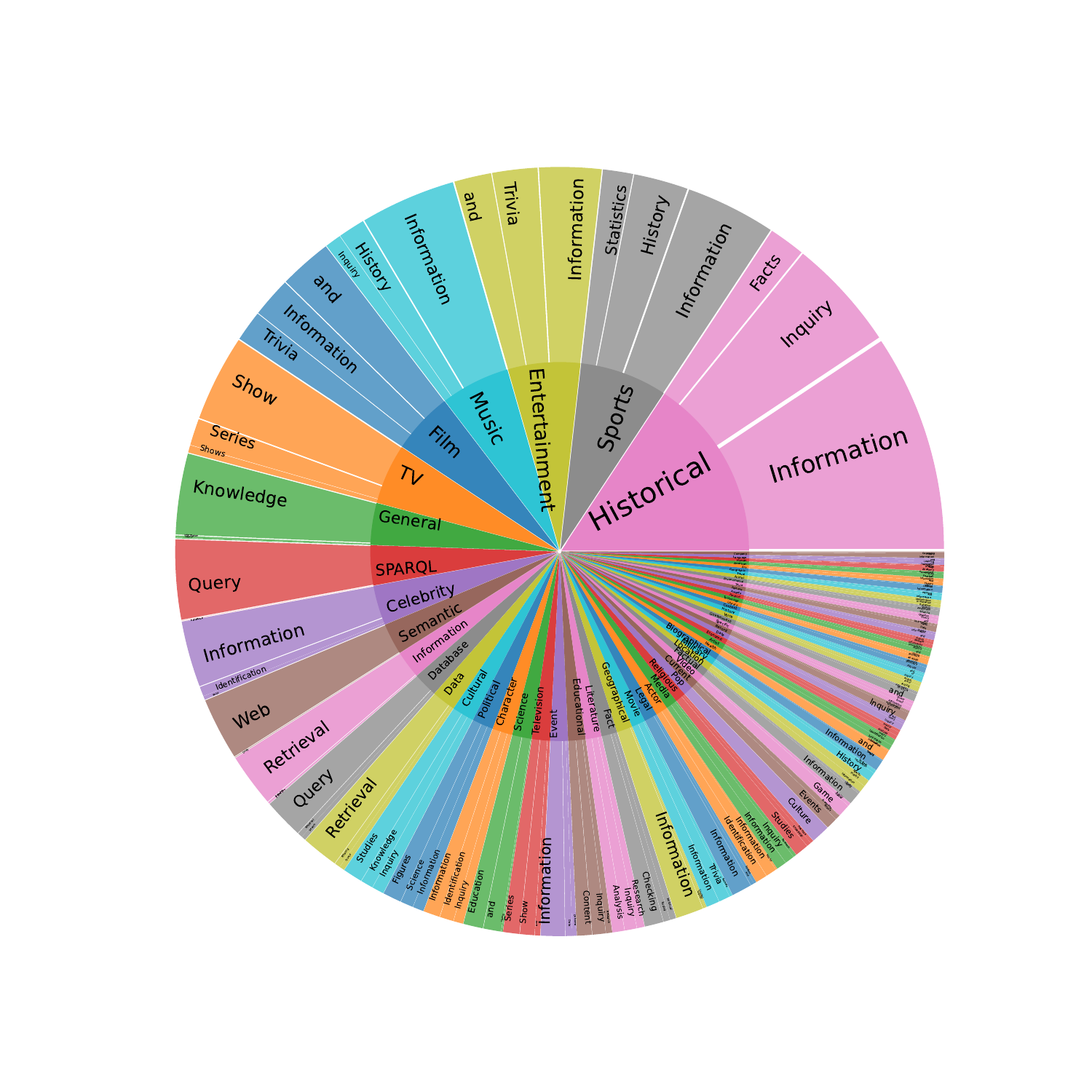}
\caption{The sunburst chart displays all tags, with each segment representing the first two words of each tag. The size of each segment is proportional to the tag's frequency.}

\label{fig:tags_visual}
\end{figure}

\paragraph{Tag Review of Training Data.}
In section ``Discussion on Query Augmentations'', we initially explore how the performance outcome is linked to complexity and diversity within the Natural Questions (NQ) dataset. Following the Instag~\citep{lu2023instag}, we also carry out an review of the intent tags within the training dataset. We randomly selected 10,000 samples from the final Supervised Fine-Tuning (SFT) data pool, which includes both the original data and 5 sets of augmented data. Figure \ref{fig:tags_visual} displays the most common tags, which predominantly pertain to historical information, sports-related data, and entertainment queries. The tags are represented by the initial two words, and their size is directly proportional to their frequency. We limit our visualization to only those tags that appear more than 600 times within our dataset.

\clearpage

\subsection{Case Studies for Preference Alignment}
To gain a deeper understanding of how DPA-RAG aligns the preferences of large models to knowledge, we conducted a case study and manual analysis, marking key supporting knowledge and misleading knowledge in the TOP-3 documents with green and red, respectively.

\begin{tcolorbox}[
colback=white!10!white,
colframe=black!75!black,
title=\large NQ-Case1,
breakable]

\section*{Baseline\centering\\[2ex]}

\textbf{Query}: Where was the diary of a wimpy kid filmed ?\\

\textbf{Reference Documents}: \\

\textbf{document1:}\\
Title: Diary of a Wimpy Kid: Rodrick Rules (film)\\
Content: The film was released on March 25, 2011. Talks of a sequel were announced after the release of the first, but was not officially announced until May 12, 2010, announcing that it would be released March 25, 2011. \textcolor{red}{Filming took place in Vancouver, British Columbia and New Westminster, British Columbia from August 23 to October 27, 2010.} The mall scene was filmed at Park Royal Mall in West Vancouver. Director Thor Freudenthal was replaced by director David Bowers (`Flushed Away'' and ``Astro Boy''). The film's trailer was shown with ``Gulliver's Travels'' on December 25, 2010. It was later online on\\
\textbf{document2:}\\
Title: Diary of a Wimpy Kid: Dog Days (film)\\
Content: The outdoor pool scenes at the beginning and the end of the movie were filmed at Steveston Outdoor pool in Richmond, BC. Filming occurred at the Steveston Outdoor pool during the beginning of September 2011. The Chinese Bunkhouse at the Steveston Shipyards in Richmond, BC was the location of the Wilderness Explorers cabin for Troop 133. During filming, stars Zachary Gordon and Robert Capron were seen riding the Corkscrew at Playland at Vancouver's fair, the PNE. A poster was leaked in March 2012. A teaser trailer was attached to ``The Three Stooges''. An advance screening for the film was held on\\
\textbf{document3:}\\
Title: Diary of a Wimpy Kid (film series)\\
Content: The ``Diary of a Wimpy Kid'' film series was released on March 25, 2011 and is based on the second book, ``Rodrick Rules'' with scenes from ``The Last Straw''. \textcolor{deepgreen}{Principal photography began on August 23, 2010, and was completed on October 27, 2010, with filming taking place in Vancouver and New Westminster. }``Rodrick Rules'' was directed by David Bowers, with Zachary Gordon reprising his role as Greg Heffley. New main characters include Holly Hills (Peyton List), Grandpa (Terence Kelly), and Bill Walter (Fran Kranz). Edward Shearmur composed the original score for the film. ``Diary of a Wimpy Kid: Dog Days'' is the third film in the series.\\

\textbf{Output}: \textcolor{red}{New Westminster, British Columbia} \textcolor{red}{\ding{53}} \\

\noindent
***************************************************************************
\section*{DPA-RAG\centering\\[2ex]}

\textbf{Query}: Where was the diary of a wimpy kid filmed ?\\

\textbf{Reference Documents}: \\

\textbf{document1:}\\
Title: Diary of a Wimpy Kid: Dog Days (film)\\
Content:
Outdoor pool scenes at the beginning and the end of the movie were filmed at Steveston Outdoor pool in Richmond, B.C. Filming occurred at the Steveston Outdoor pool during the beginning of September 2011. The Chinese Bunkhouse at the Steveston Shipyards in Richmond, B.C. was the location of the Wilderness Explorers cabin for Troop 133. During filming, stars Zachary Gordon and Robert Capron were seen riding the Corkscrew at Playland at Vancouver's fair, the P.N.E. A poster was leaked in March 2012. A teaser trailer was attached to ``The Three Stooges''. An advance screening for the film was held on...\\

\textbf{document1:}\\
Title: Diary of a Wimpy Kid: Dog Days (film)\\
Content:
All hang out together at the local pool. Greg describes that his summer did not go the way he expected, but had positive circumstances. This film is the last movie in the ``Diary of a Wimpy Kid'' film series to feature the original cast, as they aged out of their roles as middle-schoolers. Principal photography began on August 8, 2011, in Vancouver and was completed on October 7, 2011. The location for the country club pool was Eagle Ridge Outdoor pool in Coquitlam, B.C. Filming at Eagle Ridge Outdoor pool took place during the end of August 2011. The municipal...\\

Title: Diary of a Wimpy Kid (film series)\\
Content:
``Diary of a Wimpy Kid'' film series. It was released on March 25, 2011, and is based on the second book, ``Rodrick Rules'' with scenes from ``The Last Straw''. \textcolor{deepgreen}{Principal photography began on August 23, 2010, and was completed on October 27, 2010, with filming taking place in Vancouver and New Westminster.} ``Rodrick Rules'' was directed by David Bowers, with Zachary Gordon reprising his role as Greg Heffley. New main characters include Holly Hills (Peyton List), Grandpa (Terence Kelly), and Bill Walter (Fran Kranz). Edward Shearmur composes the original score for the film. ``Diary of a Wimpy Kid: Dog Days'' is the third film...\\

\textbf{Output}: {\color{deepgreen}Vancouver} $\color{deepgreen}\checkmark $\\
\\
\noindent
***************************************************************************\\

\textbf{Analysis}: The retrieved documents of the baseline contain both aligned knowledge and unaligned knowledge, with the final reasoning being misled by the unaligned knowledge. DPA-RAG filters out the unaligned knowledge during the Reranker process, retaining only the aligned knowledge, leading to successful reasoning in the end.\\

\label{tab:NQ-Case2}
\end{tcolorbox}

\clearpage

\begin{tcolorbox}[
colback=white!10!white,
colframe=black!75!black,
title=\large NQ-Case2,
breakable]

\section*{Baseline\centering\\[2ex]}

\textbf{Query}: Three largest cities in the world by population? \\

\textbf{Reference Documents}: \\

\textbf{document1:}\\
Title: Americas\\
Content: Three public bodies of the Netherlands, two unincorporated territories of the United States, and one uninhabited territory of the United States. In 2015 the total population of the Americas was about 985 million people, divided as follows: There are three urban centers that each hold titles for being the largest population area based on the three main demographic concepts: In accordance with these definitions, the three largest population centers in the Americas are: Mexico City, anchor to the largest metropolitan area in the Americas; New York City, anchor to the largest urban area in the Americas; and São Paulo, the... \\

\textbf{document2:}\\
Title: Europe\\
Content: Are recognised political goals in Europe today. The Council of Europe Framework Convention for the Protection of National Minorities and the Council of Europe's European Charter for Regional or Minority Languages set up a legal framework for language rights in Europe. The four most populous cities of Europe are Istanbul, Moscow, Paris and London, each have over 10 million residents, and as such have been described as megacities. \textcolor{red}{While Istanbul has the highest total population, one third lies on the Asian side of the Bosporus, making Moscow the most populous city entirely in Europe.} The next largest cities in order... \\

\textbf{document3:}\\
Title: World population\\

Content: Permanently inhabited on a large scale. Asia is the most populous continent, with its 4.54 billion inhabitants accounting for 60\% of the world population. The world's two most populated countries, China and India, together constitute about 36\% of the world's population. Africa is the second most populated continent, with around 1.28 billion people, or 16\% of the world's population. Europe's 742 million people make up 10\% of the world's population as of 2018, while the Latin American and Caribbean regions are home to around 651 million (9\%). Northern America, primarily consisting of the United States and Canada, has a population...\\

\textbf{Output}: \textcolor{red}{Istanbul} \textcolor{red}{\ding{53}} \\

\noindent
***************************************************************************
\section*{DPA-RAG\centering\\[2ex]}

\textbf{Query}: Three largest cities in the world by population? \\

\textbf{Reference Documents}: \\

\textbf{document1:}\\
Title: Beijing\\
Content: Resided in urban districts or suburban townships, and 2.897 million lived in rural villages. The encompassing metropolitan area was estimated by the OECD (Organisation for Economic Co-operation and Development) to have, a population of 24.9 million. \textcolor{deepgreen}{Within China, the city ranked second in urban population after Shanghai and the third in municipal population after Shanghai and Chongqing. Beijing also ranks among the most populous cities in the world, a distinction the city has held for much of the past 800 years, especially during the 15th to early 19th centuries when it was the largest city in the world. About...} \\

\textbf{document2:}\\
Title: City\\
Content: A fifth of the population is said to live in shantytowns (favelas, poblaciones callampas, etc.). Batam, Indonesia, Mogadishu, Somalia, Xiamen, China, and Niamey, Niger, are considered among the world's fastest-growing cities, with annual growth rates of 5–8\%. In general, the more developed countries of the ``Global North'' remain more urbanized than the less developed countries of the ``Global South''—but the difference continues to shrink because urbanization is happening faster in the latter group. Asia is home to by far the greatest absolute number of city-dwellers: over two billion and counting. The UN predicts an additional 2.5 billion city dwellers ...\\

\textbf{document3:}\\
Title: Russia\\
Content: Russia, officially the Russian Federation, is a country in Eurasia. At, Russia is the largest country in the world by area, covering more than one-eighth of the Earth's inhabited land area, and the ninth most populous, with about 144.5 million people, excluding Crimea. About 77\% of the population live in the western, European part of the country. Russia's capital, Moscow, is the largest metropolitan area in Europe proper and one of the largest cities in the world; other major cities include Saint Petersburg, Novosibirsk, Yekaterinburg, and Nizhny Novgorod. Extending across the entirety of Northern Asia...\\

\textbf{Output}: {\color{deepgreen}Shanghai} $\color{deepgreen}\checkmark $
\\
\noindent
***************************************************************************\\

\textbf{Analysis}: The retrieved documents of the baseline only contain unaligned knowledge, with the final reasoning being misled by the unaligned knowledge. DPA-RAG filters out the unaligned knowledge during the reranking process. Moreover, it also find the aligned knowledge from documents with lower vector similarity, reranking them to the front. Finally, the aligned knowledge leads to successful reasoning in the end.\\

\label{tab:TQA-Case1}
\end{tcolorbox}

\clearpage

\begin{tcolorbox}[
colback=white!10!white,
colframe=black!75!black,
title=\large TQA-Case1,
breakable]

\section*{Baseline\centering\\[2ex]}

\textbf{Query}: Which volcano in Tanzania is the highest mountain in Africa? \\

\textbf{Reference Documents}: \\

\textbf{document1:}\\
Title: Mount Kilimanjaro\\
Content: Mount Kilimanjaro or just Kilimanjaro, with its three volcanic cones, ``Kibo,'' ``Mawenzi,'' and ``Shira,'' is a dormant volcano in Tanzania. It is the highest mountain in Africa, about [height missing] from its base, and [elevation missing] above sea level. The first people known to have reached the summit of the mountain were Hans Meyer and Ludwig Purtscheller in 1889. The mountain is part of the Kilimanjaro National Park and is a major climbing destination. The mountain has been the subject of many scientific studies because of its shrinking glaciers and disappearing ice fields. Kilimanjaro is a large stratovolcano and is composed...\\

\textbf{document2:}\\
Title: Mount Kilimanjaro\\
Content: \textcolor{red}{Of three distinct volcanic cones: Kibo, the highest; Mawenzi at; and Shira, the shortest at. Mawenzi and Shira are extinct, while Kibo is dormant and could erupt again.} Uhuru Peak is the highest summit on Kibo's crater rim. The Tanzania National Parks Authority, a Tanzanian governmental agency, and the United Nations Educational, Scientific and Cultural Organization list the height of Uhuru Peak as [height missing]. That height is based on a British Ordnance Survey in 1952. Since then, the height has been measured as [height missing] in 1999, [height missing] in 2008, and [height missing] in 2014. The interior of the volcanic edifice is poorly...\\

\textbf{document3:}\\
Title: Mount Kilimanjaro\\
Content: \textcolor{deepgreen}{Mount Kilimanjaro or just Kilimanjaro, with its three volcanic cones, ``Kibo,'' ``Mawenzi,'' and ``Shira,'' is a dormant volcano in Tanzania. It is the highest mountain in Africa, about ; from its base, and  above sea level.} The first people known to have reached the summit of the mountain were Hans Meyer and Ludwig Purtscheller in 1889. The mountain is part of...\\

\textbf{Output}: \textcolor{red}{Mawenzi} \textcolor{red}{\ding{53}} \\

\noindent
***************************************************************************
\section*{DPA-RAG\centering\\[2ex]}

\textbf{Query}: Which volcano in Tanzania is the highest mountain in Africa? \\

\textbf{Reference Documents}: \\

\textbf{document1:}\\
Title: Mount Kilimanjaro\\
Content: \textcolor{deepgreen}{Mount Kilimanjaro or just Kilimanjaro, with its three volcanic cones, ``Kibo,'' ``Mawenzi,'' and ``Shira,'' is a dormant volcano in Tanzania. It is the highest mountain in Africa, about ; from its base, and  above sea level.} The first people known to have reached the summit of the mountain were Hans Meyer and Ludwig Purtscheller in 1889. The mountain is part of...\\

\textbf{document2:}\\
Title: Mount Kilimanjaro\\
Content: Mount Kilimanjaro or just Kilimanjaro, with its three volcanic cones, ``Kibo,'' ``Mawenzi,'' and ``Shira,'' is a dormant volcano in Tanzania. It is the highest mountain in Africa, [height missing] from its base, and [altitude missing] above sea level. The first people known to have reached the summit of the mountain were Hans Meyer and Ludwig Purtscheller in 1889. The mountain is part of the Kilimanjaro National Park and is a major climbing destination. The mountain has been the subject of many scientific studies because of its shrinking glaciers and disappearing ice fields. Kilimanjaro is a large stratovolcano and is composed...\\

\textbf{document3:}\\
Title: Geography of Tanzania\\
Content: Geography of Tanzania Tanzania comprises many lakes, national parks, and Africa's highest point, Mount Kilimanjaro [altitude missing]. Northeast Tanzania is mountainous and includes Mount Meru, an active volcano, Mount Kilimanjaro, a dormant volcano, and the Usambara and Pare mountain ranges. Kilimanjaro attracts thousands of tourists each year. West of those mountains is the Gregory Rift, which is the eastern arm of the Great Rift Valley. On the floor of the rift are a number of large salt lakes, including Natron in the north, Manyara in the south, and Eyasi in the southwest. The rift also encompasses the Crater Highlands, which includes...\\

\textbf{Output}: {\color{deepgreen}Kilimanjaro} $\color{deepgreen}\checkmark $
\\
\noindent
***************************************************************************\\

\textbf{Analysis}: The baseline's retrieved documents include both aligned and unaligned knowledge, with unaligned knowledge ranked higher due to its similarity, leading to incorrect inferences. Conversely, DPA-RAG sorted aligned knowledge to top-1 and filtered out unaligned knowledge in reranking process, ultimately succeeding in inference.\\

\label{tab:TQA-Case2}
\end{tcolorbox}

\clearpage

\begin{tcolorbox}[
colback=white!10!white,
colframe=black!75!black,
title=\large TQA-Case2,
breakable]

\section*{Baseline\centering\\[2ex]}

\textbf{Query}: What nationality was the painter Vincent van Gogh?\\

\textbf{Reference Documents}: \\

\textbf{document1:}\\
Title: Vincent van Gogh\\
Content: Vincent Willem van Gogh (30 March 1853–29 July 1890) was a Dutch Post-Impressionist painter who is among the most famous and influential figures in the history of Western art. In just over a decade, he created about 2,100 artworks, including around 860 oil paintings, most of them in the last two years of his life. They include landscapes, still lifes, portraits, and self-portraits, and are characterized by bold colors and dramatic, impulsive, and expressive brushwork that contributed to the foundations of modern art. However, he was not commercially successful, and his suicide at 37 followed years...

\textbf{document2:}\\
Title: Theo van Gogh (art dealer)\\
Content: Theodorus ``Theo'' van Gogh was born on 1 May 1857 in the village Groot-Zundert in the province of North Brabant, Netherlands. He was the son of Theodorus van Gogh and Anna Cornelia Carbentus. His elder brother was artist Vincent van Gogh (1853–1890). Theo worked for some years at the Dutch office of the Parisian art dealers Goupil \& Cie in The Hague. Theo joined the Brussels office on 1 January 1873 as their youngest employee. After Theo was transferred to the London office, he moved to the office in...\\

\textbf{document3:}\\
Title: Vincent van Gogh\\
Content: Van Gogh Museum opened in the Museumplein in Amsterdam in 1973. It became the second most popular museum in the Netherlands, after the Rijksmuseum, regularly receiving more than 1.5 million visitors a year. In 2015 it had a record 1.9 million; 85 percent of the visitors come from other countries. \textcolor{deepgreen}{Vincent Willem van Gogh (30 March 1853–29 July 1890) was a Dutch Post-Impressionist painter who is among the most famous and influential figures in the history of Western art.} In just over a decade, he created about 2,100 artworks, including around 860 oil paintings, most of...\\

\textbf{Output}: \textcolor{red}{Autochtones} \textcolor{red}{\ding{53}} \\

\noindent
***************************************************************************
\section*{DPA-RAG\centering\\[2ex]}

\textbf{Query}: Three largest cities in the world by population? \\

\textbf{Reference Documents}: \\

\textbf{document1:}\\
Title: Vincent van Gogh\\
Content: The Van Gogh Museum opened in the Museumplein in Amsterdam in 1973. It became the second most popular museum in the Netherlands, after the Rijksmuseum, regularly receiving more than 1.5 million visitors a year. In 2015, it had a record 1.9 million; 85 percent of the visitors come from other countries. \textcolor{deepgreen}{Vincent Willem van Gogh (30 March 1853 – 29 July 1890) was a Dutch Post-Impressionist painter who is among the most famous and influential figures in the history of Western art.} In just over a decade, he created about 2,100 artworks, including around 860 oil paintings, most of...\\

\textbf{document2:}\\
Title: Vincent van Gogh\\
Content: \textcolor{deepgreen}{Vincent Willem van Gogh (30 March 1853 – 29 July 1890) was a Dutch Post-Impressionist painter who is among the most famous and influential figures in the history of Western art.} In just over a decade, he created about 2,100 artworks, including around 860 oil paintings, most of them in the last two years of his life. They include landscapes, still lifes, portraits, and self-portraits, and are characterised by bold colours and dramatic, impulsive and expressive brushwork that contributed to the foundations of modern art. However, he was not commercially successful, and his suicide at 37 followed years...\\

\textbf{document3:}\\
Title: Vincent van Gogh\\
Content: Vincent Willem van Gogh was born on 30 March 1853 into a Dutch Reformed family in Groot-Zundert, in the predominantly Catholic province of North Brabant in the southern Netherlands. He was the oldest surviving child of Theodorus van Gogh, a minister of the Dutch Reformed Church, and Anna Cornelia Carbentus. Van Gogh was given the name of his grandfather, and of a brother stillborn exactly a year before his birth. Vincent was a common name in the Van Gogh family: his grandfather, Vincent (1789 – 1874), who received a degree in theology at the University of Leiden in 1811, had six...\\

\textbf{Output}: {\color{deepgreen}Dtuch} $\color{deepgreen}\checkmark $\\

\noindent
***************************************************************************\\

\textbf{Analysis}: Although the baseline's retrieved documents only contains aligned knowledge, the quantity and order of relevant knowledge are relatively low. DPA-RAG not only sorts multiple aligned pieces of knowledge to the front during the reranking process, but also relies on the key information capture ability brought by the pre-aligned stage, allowing the LLMs to better focus on knowledge beneficial to inference, ultimately leading to successful reasoning.

\end{tcolorbox}

\end{document}